\documentclass[a4paper,fleqn]{cas-dc}

\usepackage[numbers]{natbib}
\usepackage{amsmath,amssymb,amsfonts}
\usepackage{algorithmic}
\usepackage[linesnumbered,algoruled,boxed,lined]{algorithm2e}
\usepackage{comment}

\usepackage[normalem]{ulem}

\usepackage{graphicx}
\usepackage{wrapfig}
\usepackage{caption}
\usepackage{subcaption}
\usepackage{textcomp}
\usepackage{xcolor}
\usepackage[inline]{enumitem}
\usepackage{adjustbox}
\usepackage{multirow, booktabs}
\usepackage{tabularray}
\usepackage{hyperref}
\usepackage{longtable}
\usepackage{lscape}

\newrobustcmd{\B}{\bfseries}
\newrobustcmd{\U}[1]{\uline{#1}}

\definecolor{orange}{HTML}{C2570E}
\definecolor{bluet}{HTML}{294483}
\definecolor{green}{HTML}{298323}
\definecolor{red}{HTML}{A31313}

\def\tsc#1{\csdef{#1}{\textsc{\lowercase{#1}}\xspace}}
\tsc{WGM}
\tsc{QE}
\begin{document}
\let\WriteBookmarks\relax
\def\floatpagepagefraction{1}
\def\textpagefraction{.001}

\shorttitle{Evaluation of Video-Based rPPG in Challenging Environments}    

%\shortauthors{<short author list for running head>}  

\title [mode = title]{Evaluation of Video-Based rPPG in Challenging Environments: Artifact Mitigation and Network Resilience}  

\tnotemark[1] 

\tnotetext[1]{This research has been supported by the Academy of Finland 6G Flagship program under Grant 346208 and PROFI5 HiDyn under Grant 32629 and JSPS (Japan Society for the Promotion of Science) KAKENHI Grant Number 21J22170.} 


\author[1]{Nhi Nguyen}[orcid=0009-0002-2090-0746]

%\cormark[<corr mark no>]

%\fnmark[<footnote mark no>]

\ead{thi.tn.nguyen@oulu.fi}

%\ead[url]{<URL>}

\credit{Writing – original draft, Writing – review \& editing, Visualization, Validation, Software, Methodology, Investigation, Formal analysis, Conceptualization}

\author[1]{Le Nguyen}
\ead{le.nguyen@oulu.fi}
\credit{Writing – original draft, Writing – review \& editing, Methodology, Investigation, Formal analysis, Conceptualization, Supervision}

\author[1,2]{Honghan Li}
\ead{lihonghan@mbm.me.es.osaka-u.ac.jp}
\credit{Writing – original draft, Visualization, Software, Methodology}

\author[1,3]{Miguel \ Bordallo\ López}
\ead{miguel.bordallo@oulu.fi}
\credit{Writing – original draft, Writing – review \& editing, Methodology, Investigation, Conceptualization, Funding acquisition}

\author[1]{Constantino \'{A}lvarez\ Casado}
\ead{constantino.alvarezcasado@oulu.fi}
\credit{Writing – original draft, Writing – review \& editing, Visualization, Validation, Software, Methodology, Investigation, Formal analysis, Conceptualization, Supervision}

%\cortext[1]{Corresponding author}



\affiliation[1]{organization={Center for Machine Vision and Signal Analysis (CMVS), University of Oulu},
            city={Oulu},
            country={Finland}}
            
\affiliation[2]{organization={Division of Bioengineering, Graduate School of Engineering Science, Osaka University},
            city={Osaka},
            country={Japan}}
\affiliation[3]{organization={VTT Technical Research Center of Finland Ltd.},
            city={Oulu},
            country={Finland}}

% TINO NOTES TO REMEMBER:
% We have to comment that we present an evaluation framework.

\begin{abstract}
Video-based remote photoplethysmography (rPPG) has emerged as a promising technology for non-contact vital sign monitoring, especially under controlled conditions. However, the accurate measurement of vital signs in real-world scenarios faces several challenges, including artifacts induced by videocodecs, low-light noise, degradation, low dynamic range, occlusions, and hardware and network constraints. In this article, we systematically investigate comprehensive investigate these issues, measuring their detrimental effects on the quality of rPPG measurements. Additionally, we propose practical strategies for mitigating these challenges to improve the dependability and resilience of video-based rPPG systems. We detail methods for effective biosignal recovery in the presence of network limitations and present denoising and inpainting techniques aimed at preserving video frame integrity. Through extensive evaluations and direct comparisons, we demonstrate the effectiveness of the approaches in enhancing rPPG measurements under challenging environments, contributing to the development of more reliable and effective remote vital sign monitoring technologies.
\end{abstract}


%\item 
%\item 
%\item 

% Keywords

\maketitle

% Main text
\section{Introduction}
    Remote photoplethysmography (rPPG) marks a significant advancement in non-contact vital sign monitoring \cite{huang2023challenges}. This approach, leveraging video-based techniques, enables the measurement of cardiovascular signals without requiring direct skin contact \cite{Face2PPGPipeline2022,pyVHR2020}. Particularly significant in the context of medical diagnostics, remote healthcare, and human-computer interaction, rPPG technology offers a non-intrusive and scalable solution for heart rate monitoring. However, the practical implementation of video-based rPPG presents substantial challenges \cite{Curran2023rPPGPerspectives,huang2023challenges}. These challenges arise from a variety of factors inherent in video capture, but also in video data transmission. Major challenges in video-based rPPG include video codec distortions and image artifacts, which can affect the quality of the video signal and thereby the accuracy of rPPG measurements \cite{huang2023challenges,Rapczynski2019VideoEnco}. Issues such as noise and signal degradation, often due to poor recording conditions (low-light conditions) or problems during transmission, often accentuate these challenges \cite{Williams2023IlluminationrPPGs,alvarez2023assessingRPPGsVideoConf}. Limitations in the dynamic range of cameras and video codecs complicate the detection of subtle skin color changes essential for accurate rPPG analysis \cite{Rapczynski2019VideoEnco,McDuff2017VideoCodecs}. Occlusions in video frames, such as those caused by movements, obstructive objects, facial accessories, or privacy anonymization methods can interrupt the extraction of blood volume pulse (BVP) signals from the face \cite{Nhi2023Deteriorated}. Lastly, hardware and network constraints, such as limited processing power and bandwidth, can degrade the performance of real-time data processing and transmission \cite{alvarez2023assessingRPPGsVideoConf}, crucial for applications like telemedicine and remote patient monitoring. These issues compromise the accuracy of vital sign measurements and limit the adaptability of rPPG in real-world scenarios. Addressing these challenges is imperative for advancing and broadening the use of rPPG technology \cite{Curran2023rPPGPerspectives,huang2023challenges}.

%%This is particularly significant in an era where remote health monitoring and telehealth are becoming increasingly prevalent, driven by factors like the aging global population, the rise in chronic health conditions, geographical barriers, lack of healthcare personnel and the need for healthcare systems to maximize efficiency and reach \cite{Smith2005TelemedicineRural,SEKHON2021DementiaTele,xiao2023TeleChronic}.

%In our previous publications, we have explored various aspects of these challenges. From examining the impact of network and computing constraints on rPPG via videocalls \cite{alvarez2023assessingRPPGsVideoConf} to analyzing the performance of rPPG methods on videos with deliberately deteriorated quality \cite{Nhi2023Deteriorated}, our research has consistently aimed to enhance the reliability and applicability of rPPG technology. Building on this foundation, our current paper extends the scope of our results to address a wider array of challenges in video-based rPPG. We specifically focus on spatial artifacts, temporal artifacts, and visual occlusions, as well as offering new solutions to these complex problems. 

This article describes an overview of rPPG technology, reviewing related research and outlining challenges in video-based applications. We introduce a systematic evaluation framework to assess how these challenges affect the quality of rPPG systems. Alongside this, we propose and validate potential solutions to tackle these challenges through extensive testing on public databases. The paper concludes by discussing the implications of our findings, emphasizing the potential for improved remote vital sign monitoring, thereby providing engineers and researchers with informed insights for system design and integration into healthcare solutions. Our contributions include:

\begin{itemize}

    \item A comprehensive evaluation of rPPG signals extracted using both learning-based (specifically, deep learning) and non-learning-based (predominantly computer vision and signal processing) methods. This comparison assesses the effectiveness of each approach in managing artifacts, aiding in the identification of robust signal enhancement strategies.
    
    \item Quantifying the impact of spatial, temporal, and visual occlusion artifacts on rPPG signal quality, including intentional occlusions (anonymization), noise related to low-light conditions, color depth resolution, camera frame rate, stream resolution, and the impact of connectivity or computing resource limitations leading to random frame dropping.

    \item Proposing and evaluating of denoising and inpainting strategies to improve rPPG accuracy in measuring heart rate in noisy and occluded conditions.
    
    \item A comparative evaluation of mitigation strategies to improve rPPG signals in the presence of computation and network-related issues.
    
\end{itemize}

We evaluate these issues and mitigation strategies across a broad selection of seven publicly available rPPG databases, including UBFC-rPPG 1 \& 2, UCLA-rPPG, UBFC-Phys, PURE, MAHNOB, LGI-PPGi, and COHFACE. This comprehensive evaluation encompasses a diverse range of scenarios, from typical static and resting scenarios to physical activities in gym settings or video call interactions.

    \label{sec:introduction}

\section{Related Work}
Remote photoplethysmography (rPPG) is a non-invasive, contactless method that uses a video camera and ambient light to detect physiological signals by observing variations in skin color due to blood flow, reflecting the complex interactions of light with skin properties\cite{huang2023challenges,rPPGFundaments2015}. As light hits the skin, it reflects based on the angle of incidence and skin characteristics, including polarization, and then penetrates the skin, scattering when meeting skin structures like cells and blood vessels. This process is influenced by skin chromophores like melanin and hemoglobin, which absorb specific light wavelengths. These changes in light intensity, recorded by the camera, correlate with the cardiac cycle; more light is absorbed during the systole when blood volume increases, and less during diastole. To improve signal detection accuracy, rPPG often employs green or near-infrared wavelengths, which hemoglobin preferentially absorbs.

However, the rPPG technology faces several intrinsic challenges. These are challenges inherent to the BVP signal capture process, tied to the immediate measurement environment and subject-specific factors such as motion artifacts due to subject movement \cite{Face2PPGPipeline2022}, distance from the camera sensor to the skin area and resolution of the facial region\cite{Song2020rPPGResolutionDistance}, variations in skin pigmentation \cite{Setchfield2024TypeSkin}, environmental light conditions \cite{Lin2017IlluminationColor}, or occlusions \cite{Nhi2023Deteriorated}, either from external factors or from purposeful anonymization techniques. All these factors can compromise the heart rate estimations \cite{huang2023challenges}. Furthermore, extrinsic challenges are external conditions and technical limitations, mostly stemming from network and computing constraints that can significantly impact the performance of rPPG methods if not adequately addressed during development. These challenges, arising during video capture, processing, and transmission, include bandwidth limitations, packet loss, latency, video compression artifacts, frame rate variability, camera sensor's rolling shutter effects, and restricted computational resources. Such factors can substantially degrade video quality, affecting the detection of subtle skin color variations crucial for accurate rPPG signal extraction and subsequent analysis \cite{alvarez2023assessingRPPGsVideoConf,Mironenko2020RareFactorsRPPGs}.

\subsection{Intrinsic challenges}

To address these challenges, studies have concentrated on enhancing rPPG's resilience to motion, adapting to skin tone variations, or improving signal quality amid noise and changes in illumination \cite{Face2PPGPipeline2022}. Traditional non-learning-based methodologies in rPPG have been extensively directed toward addressing physical artifacts by employing sophisticated models of light reflection, signal processing techniques, and blood volume pulse (BVP) extraction from various color channels. Research efforts have largely aimed at isolating BVP signals from raw RGB video data using a variety of mathematical models and algorithms. Notably, techniques such as Plane Orthogonal to Skin (POS), Chrominance-based (CHROM), Orthogonal to Motion and Intensity (OMIT), Local Group Invariance (LGI), Chrominace-KDICA (CK), Blind Source Separation (BSS), and Independent Component Analysis (ICA) have been developed to isolate the pulsatile component of blood volume from the skin reflections recorded in RGB channels \cite{Face2PPGPipeline2022,pyVHR2020,Haugg2022rPPGEffectiveness}. Among the earliest approaches, the GREEN method \cite{GreenMethod2008}, proposed in 2008, utilizes the green video channel, which has been found to capture a stronger pulsatile signal relative to the red and blue channels \cite{ontiveros14evaluating}. To address motion artifacts, subsequent research has introduced strategies for stabilizing the region of interest, particularly focused in faces, using optical flow techniques and tracking algorithms \cite{Kumar2020PulseCam,boccignone2022pyvhr}. Efforts to counteract the effects of head and facial movements have led to the development of face tracking and alignment approaches \cite{Face2PPGPipeline2022}. Additionally, several filtering approaches, including bandpass, adaptive, detrending, and LSTM-based filters, have been utilized to clean PPG signals from noise and motion-related disturbances \cite{boccignone2022pyvhr,Kumar2022RobustPPG,Botina2020LSTMfilter}. Signal separation methods based on independent component analysis or blind source separation, such as PCA \cite{PCAMethod2011}, CK \cite{Song2020rPPGResolutionDistance}, OMIT \cite{Face2PPGPipeline2022} or ICA \cite{PohMethod2011}, have been utilized to isolate the genuine physiological signals from uncorrelated signals and noise. Reflective models like CHROM \cite{CHROMMethod2013} and POS \cite{POSMethod2017} have been developed to separate the specular and diffuse reflection components, which contain the pulsatile physiological signals. Approaches like \cite{LiMethod2014} and \cite{Wang2019rPPGSignature} have focused on correlating signals using normalized reference waveforms or noise reference signals, while others have concentrated on addressing illumination variations \cite{Nowara2021Denoising,LiMethod2014} and skin tone variations \cite{Kumar2015DistancePPG} to enhance the signal quality further. Exploring RGB channels further, the GRGB rPPG method \cite{Haugg2023GRGBrPPG} utilizes green-to-red (GR) and green-to-blue (GB) ratios, and their sum (GR + GB), to improve rPPG signal quality efficiently.

Deep Learning and other learning-based approaches in rPPG have emerged in recent years to tackle intrinsic artifacts by learning physical reflective and absorption models for BVP in a data-driven manner. These methods offer automated solutions for issues like noise, movements, illumination changes, or video quality, enhancing heart rate estimation accuracy. Key developments include HR-CNN \cite{HRCNNMethod2018}, an end-to-end two-step CNN, comprised of an extractor trained to optimize SNR for rPPG signal extraction from video frames, and an HR estimator that minimizes the mean absolute error (MAE) between predicted and ground-truth HR values, and DeepPhys \cite{DeepPhysMethod2018}, which utilizes deep networks with attention mechanisms and skin reflection models for signal extraction. RhythmNet \cite{RhythmNet2019} employs spatial-temporal mapping for continuous HR measurement, while \cite{ZitongMethod2019} introduces a two-stage approach, combining video enhancement and 3D-CNN for rPPG signal recovery. Further innovations are seen in AutoHR \cite{AutoHR2020Zitong}, utilizing Neural Architecture Search for optimal signal extraction, and MetaRPPG \cite{MetaRPPG2020}, a transductive meta-learner for self-supervised learning. PulseGAN \cite{PulseGAN2021} represents a hybrid method combining unsupervised PPG extraction with a GAN for realistic pulse signal generation. The recent AND-rPPG \cite{ANDrPPG2022} focuses on using Action Units and Temporal Convolutional Networks for effective signal denoising. Recent developments in supervised deep learning have significantly improved physiological signal estimation. MTTS-CAN \cite{mttscan2020} is a multi-task variant, halves computational costs, and facilitates information sharing between blood volume pulse and respiration signal estimation. Additionally, PhysFormer \cite{physformer2021} offers an end-to-end video transformer architecture for remote physiological measurement, leveraging cascaded temporal difference transformer blocks and supervision in the frequency domain. EfficientPhys \cite{efficientphys2021} introduces two new neural architectures, a visual transformer and a convolutional network, marking the debut of visual transformers in camera-based physiological measurement and an end-to-end on-device neural architecture for mobile devices. In unsupervised deep learning, Gideon et al. \cite{gideon2021} pioneer label-free rPPG estimation using a contrastive framework and robust loss functions. SiNC \cite{sinc2023} leverages weak periodicity assumptions for accurate rPPG regression from unlabeled face videos, enabling camera-based vital sign measurements without ground-truth data. Contrast-Phys \cite{contrastphys2022} introduces a method trainable in unsupervised settings, leveraging spatiotemporal contrast for accurate rPPG measurement, robust against noise interference, and computationally efficient. 

Traditional non-learning-based rPPG methods excel in generalizing across different conditions without specific training data, but face challenges with noise sensitivity and varying skin types. On the other hand, learning-based methods, offer high accuracy and effective noise handling. These methods usually need minimal intervention beyond essential preprocessing, and they learn to recognize facial noise patterns in relation to ground-truth PPG signals \cite{Survey_DL_RPPG2021_Cheng}. However, their dependency on representative training data restricts their generalizability and raises concerns about data acquisition costs and computational resources. Additionally, they often function as `black-box' models with limited transparency, a significant drawback in healthcare applications\cite{CNNrPPG_Limitations2020}.

\subsection{Extrinsic challenges}

Extrinsic factors, especially those related to network and computational constraints, can significantly impact the performance of rPPG systems \cite{alvarez2023assessingRPPGsVideoConf}, but this influence is often not fully considered in research. Not many studies have focused in this aspects. In the process of designing and implementing rPPG methods, many researchers assume ideal operating conditions,  overlooking the complexities of deploying these systems in real-world environments, especially for real-time applications. This gap becomes particularly relevant in video streaming for real-time vital sign monitoring, where maintaining signal quality with low latency and limited bandwidth is crucial.

The main extrinsic factors impacting rPPG systems include video compression artifacts, frame rate variability, network jittering and resolution adjustments. These can compromise the integrity of rPPG signals. Video compression may add noise and alter the time-based patterns essential for extracting accurate BVP signals. Likewise, changes in frame rate and resolution can modify the signal's morphological and frequency attributes, affecting the accuracy of physiological signal detection across different network conditions.

Some research studies have demonstrated that video compression may induce minor degradation in signal quality, affecting the features and morphology of the rPPG signal \cite{hanfland2016videocomp,McDuff2017VideoCodecs,Rapczynski2019VideoEnco,Lampier2019ImageResolution}. To address video compression artifacts, techniques including image filtering and deep learning-based methods have been proposed \cite{hanfland2016videocomp,ZitongMethod2019}. Singular spectrum analysis for signal reconstruction and selection has also been explored \cite{Zhao2018rPPGinCompressed}. Studies on the effect of frame rate and image resolution suggest that maintaining typical values (15-200fps) does not significantly impact heart rate estimation accuracy \cite{Sun2013FrameRate,Blackford2015FrameRateReso}. Efficient PPG signal coding strategies have been studied, but these require specific learning approaches and architecture \cite{Zhao2019POSSC}. Other studies have explored the effect of different image formats in DL-based rPPG approaches \cite{botina2022rtrppg}. For example, Álvarez et al. \cite{alvarez2023assessingRPPGsVideoConf} have conducted an in-depth analysis of the effects associated with regular and random frame dropping, as well as frame jittering, phenomena typically encountered during video calls. The study proposes several strategies to mitigate their detrimental impact on signal integrity. Additionally, Nguyen et al. \cite{Nhi2023Deteriorated} investigated the impact of noise on the performance of rPPG systems, both non-learning-based and learning-based, and have proposed denoising techniques to alleviate its adverse effects. Although these are clear steps to properly characterize these effects, further investigation is needed to fully understand and address other extrinsic factors like latency, packet loss, and computing resource limitations. 

    \label{sec:relatedwork}

\section{Evaluation Framework}
\begin{figure*}[ht!]
  \begin{center}
    \includegraphics[width=\textwidth]{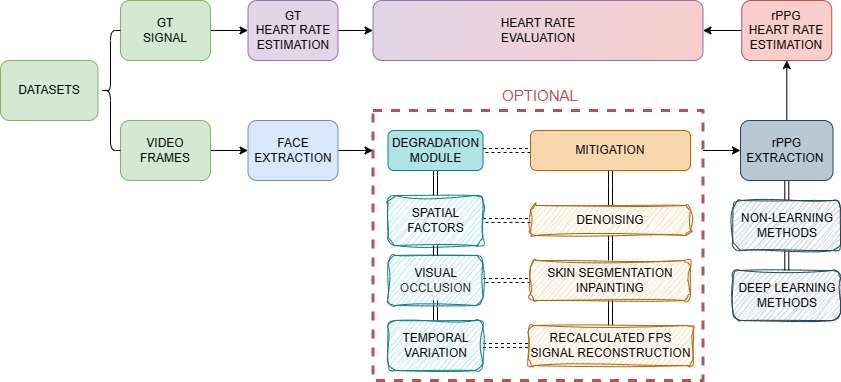}
  \end{center}
  \caption{Evaluation Framework}
  \label{fig:methodology}
\end{figure*}

    For the evaluation of rPPGs in a challenging environment, we propose a framework that integrates two optional modules into typically standard rPPG pipelines. These modules are designed to simulate potential degradation artifacts and implement corresponding mitigation strategies respectively. A simplified view of our proposed framework is depicted in Fig~\ref{fig:methodology}.

% HOMEWORKS:
% [X] Remake figure of the evaluation framework as @Miguel suggested.

Within a typical rPPG pipeline, the first step focuses on the capture and processing of video frames to detect the primary face in the input image. Subsequently, the degradation module introduces systematic spatial, temporal, and visual occlusion artifacts in the video frames. These controlled degradations aim to mimic real-world conditions, enabling the assessment of their effects on the rPPG signal extraction process. The module also allows the implementation of strategies that minimize the impact of these degradations on the performance of the rPPG techniques, improving resilience and real-world applicability. The assessment of rPPG methods may utilize either non-learning-based methods (predominantly signal processing) or learning-based approaches (mainly deep learning).

\subsection{Datasets}

The proposed evaluation framework includes seven datasets, as summarized in Table \ref{tab:dataset}. These datasets offer a wide variety of input videos, captured under different conditions, including controlled environments or natural settings, and in both compressed and uncompressed formats, for analyzing physiological signals and emotions. By examining multiple datasets, we ensure the reliability of our findings across different scenarios.

\paragraph{COHFACE \cite{heusch2017}}  contains video recordings of 40 subjects captured with a Logitech HD C525 webcam at 20 Hz, with a resolution of 640x480 pixels. The database comprises 160 videos, each lasting approximately 1 minute. Reference physiological data was recorded using medical-grade equipment.

\paragraph{LGI-PPGI \cite{pilz2018}} features 24 videos of 6 users engaged in four scenarios: gym, resting, rotation, and talking, recorded at 640×480 pixels and 25 fps using a Logitech HD C270 webcam. Reference ground-truth (GT) measurements are obtained using a CMS50E pulse oximeter device synchronized at 60 Hz.

\paragraph{MAHNOB \cite{soleymani2012}} is a multimodal database designed for emotion recognition, featuring 27 participants recorded with various cameras. The database includes 527 facial videos with corresponding reference physiological signals, recorded at 60 frames per second using an ECG sensor. In our evaluation of different experiments, only a curated subset of 81 videos from the MAHNOB-HCI database was utilized, inclusive of the first, middle, and last videos for each subject.

\paragraph{PURE \cite{stricker2014}} consists of videos of 10 subjects performing controlled head motions captured with an eco274CVGE camera at 30 Hz and a resolution of 640$\times$480 pixels. The recording session involved six distinct setups, encompassing steady movements, talking, slow translation, fast translation, slow rotation, and medium rotation. Reference pulse data was collected using a pulox CMS50E fingertip pulse oximeter at a sampling rate of 60 Hz.

\paragraph{UBFC-rPPG 1 \& 2 \cite{ubfc}} contains 50 videos synchronized with a pulse oximeter finger-clip sensor, with lengths of approximately 2 minutes. Recorded at a frame rate of 28-30Hz and a resolution of 640$\times$480 pixels in uncompressed 8-bit RGB format.

\paragraph{UCLA-rPPG \cite{ucla}} includes 98 subjects with 4-5 videos per subject, recorded at 30 fps and lasting about 1 minute each. A total of 488 videos were included, synchronized with GT heart rate data. To reduce redundancy in the dataset, only one video per subject was selected for our analysis. Specifically, the second video from each subject was used, resulting in a total of 98 videos.

\paragraph{UBFC-Phys \cite{phys}} comprises videos of 56 subjects recorded with an Edmund Optics EO-23121C RGB digital camera at 35 fps and a resolution of 1024$\times$1024 pixels. Reference GT data were collected using an Empatica E4 wristband, including BVP, skin temperature, and EDA responses. During the experiment, participants experienced social stress through a three-step process involving a resting task, a speech task, and an arithmetic task.

\begin{table}[ht!]
\setlength{\tabcolsep}{0.4em}
\def\arraystretch{1.1}
\begin{center}
\caption{Dataset Specifications}

%\resizebox{\columnwidth}{!}{

\begin{tabular}{lcccc}
\toprule
\textbf{Dataset} &\textbf{Resolution} & \textbf{Video FPS} & \begin{tabular}[c]{@{}c@{}}\textbf{Reference Signal}\\\textbf{Sampling Rate}\end{tabular} \\
\midrule
COHFACE & 640x480 & 20 & 256 \\
LGI\_PPGI & 640x480 & 25 & 60 \\
MAHNOB  & 780x580 & 61 & 256 \\
PURE & 640x480 & 30 & 60 \\
UBFC1-rPPG & 640x480 & 28 & 62 \\
UBFC2-rPPG & 640x480 & 23--29 & 30 \\
UCLA-rPPG & 640x480 & 30 & 30 \\
UBFC-Phys & 1024x1024 & 35 & 35 \\
\bottomrule
\end{tabular}
%}
\label{tab:dataset}
\end{center}
\end{table}

\subsection{Face Detection and Segmentation}
\label{sec:facedetection}
\begin{figure}[ht!]
  \begin{center}
    \includegraphics[width=0.85\columnwidth]{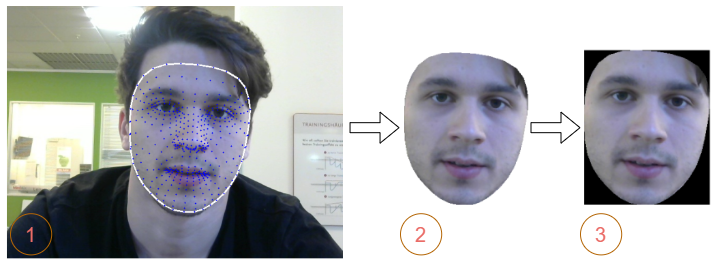}
  \end{center}
  \caption{Face extraction process in our rPPG evaluation pipeline. It involves three steps: 1) Landmarks detection, 2) Face segmentation, and 3) Background modification.}
  \label{fig:extract}
  \vspace{-3mm}
\end{figure}

In this study, we use \textit{Mediapipe} framework \cite{camillo2019}, a \textit{Google} powered machine learning solution for facial detection and segmentation from images and video streams. The detection process operates by analyzing each frame individually to detect the face, as depicted in Figure \ref{fig:extract}. 

The detection process operates by analyzing each frame individually to detect and segment the face. It consists of two real-time deep neural network models that work together: A face detector, known as BlazeFace from \textit{Google Research} \cite{bazarevsky2019blazeface}, that operates on the full image and computes face locations, and a 3D face landmark model that operates on those locations and predicts the approximate surface geometry via regression \cite{kartynnik2019real}. Using the \textit{FACEMESH\_FACE\_OVAL} mode, MediaPipe constructs a segmentation mask based on the facial outline based on 468 3D face landmarks. Subsequently, this mask is applied to the original image to segment the desired face area, which is the region of interest to extract the BVP signals. To reduce noise, especially in signal processing-based rPPG methods, the background is changed to black pixels, as shown in Figure \ref{fig:extract}. If the model fails to detect the face or its landmarks, face locations from previous frames where detection was successful are used. 

%The analysis is conducted with a maximum interval of 60 seconds in the middle of each video to address potential problems with face detection caused by sudden camera movements at the beginning or end of the video.

% To address potential problems with the face detection caused by sudden camera movements, the first and last second of the video are excluded from analysis and evaluation.

\subsection{rPPG Extraction}

We categorize rPPG extraction methods into two main groups: non-learning-based approaches and deep-learning-based techniques. We select three non-learning-based methods (NLM); OMIT~\cite{Face2PPGPipeline2022} which utilizes matrix decomposition to create an orthogonal RGB matrix and reconstructs the rPPG signal from the RGB signal using orthogonal components obtained through QR factorization; CHROM~\cite{CHROMMethod2013}, that constructs orthogonal chrominance signals to mitigate specular reflections; and POS~\cite{POSMethod2017}, that removes specular reflections using normalized RGB color space. 
In addition, we include four Deep learning methods (DLM); EfficientPhys~\cite{efficientphys2021}, that offers an end-to-end solution with a customized normalization module, tensor-shift module, and hierarchical vision transformer architecture, trained on the UBFC-rPPG dataset~\cite{ubfc}; ContrastPhys~\cite{contrastphys2022}, which employs a 3DCNN model trained with a contrastive loss function on the UBFC-rPPG dataset \cite{ubfc}; PhysFormer~\cite{physformer2021}, that utilize temporal difference transformer blocks for generating global-local rPPG features, trained on 1-fold of the VIPL-HR dataset~\cite{vipl} using a curriculum learning paradigm; and MTTS-CAN~\cite{mttscan2020}, which employs a multi-task temporal shift convolutional attention network for effective temporal modeling and signal source separation, with pre-trained models trained on the AFRL dataset~\cite{afrl}. 

% Homeworks for Nhi:
% [ ] This part needs to be more clear.

While non-learning-based methods do not restrict input length, pre-trained deep learning methods often come with specific input size requirements ($K$) to be operational. Processing an entire video, composed of ($N$) frames, as input at once for DLM is usually not possible due to limited GPU and memory resources. Therefore, most methods call for dividing the video into $M$ segments into specific lengths, where $M = \frac{N}{K}$ is advisable. If the last chunk contains a residual number of frames ($R$) smaller than the required input size $K$ for deep learning models we use frames from the previous one to pad it. In our experiments, we conducted preprocessing steps to align our data with the pre-trained data as closely as possible, using duplicate or extra frames in the input if required by the methods.  
Table \ref{tab:DLM} provides configuration details about the input and pre-trained models utilized in our study.

\begin{table*}[ht!]
\setlength{\tabcolsep}{1.5em}
\def\arraystretch{1.1}
\caption{Basic Specifications for Deep Learning Methods}
%\resizebox{\textwidth}{!}{
\begin{tabular}{llll}
\toprule
\textbf{DLM} & 
% \begin{tabular}[l]{@{}l@{}}\textbf{Pretrained}\\\textbf{Dataset}\end{tabular}  &
\textbf{Pretrained Dataset} & 
\textbf{Chunk Size} &
\textbf{Input Preprocessing}  
% \begin{tabular}[l]{@{}l@{}}\textbf{Extra}\\\textbf{Frame}\end{tabular} \\
\\
\midrule
PhysFormer \cite{physformer2021} & VIPL-HR & 160 & Image-Level Normalization \\
ContrastPhys \cite{contrastphys2022} & UBFC-rPPG & FPS$\times$10 & None \\
EfficientPhys \cite{efficientphys2021} & UBFC-rPPG & FPS$\times$10 & Video-Level Standardization \\
MTTS-CAN \cite{mttscan2020} & ARFL & 320 & \begin{tabular}[l]{@{}l@{}}Video-Level Appearance\\and Motion Calculation\end{tabular} \\
\bottomrule
\addlinespace
\end{tabular}
%}
%\scriptsize{FSD: Face Segmentation Dataset obtained from the methods described in Section \ref{sec:facedetection}.}
\label{tab:DLM}
\vspace{-3mm}
\end{table*}

\subsection{Heart Rate Estimation}

After extracting the rPPG signal utilizing several rPPG methods and using the GT signal from datasets, the signals are segmented into distinct intervals using sliding windows of 10-second and 1-second steps (hence, 9-second overlap). Each signal window undergoes filtering, both forward and backward, configured as a bandpass filter within the range of 0.75 to 4 Hz (equivalent to 45–240 bpm). 

%It is noteworthy that, exclusively for the EfficientPhys and MTTS-CAN pre-trained model methods, the cumulative sum of the rPPG signal needs to be computed before the signal undergoes filtering.

Subsequently, heart rate estimation is performed using Welch's method with a specified nFFT value of 4096. This involves applying a Hamming window to each segment, with the length of each segment matching the length of the sliding signal window ($n$). Additionally, the number of points to overlap between segments is set to $\lfloor \frac{n}{8} \rfloor$, contributing to a robust and precise estimation of heart rate.

\subsection{Evaluation Metrics}

To evaluate the accuracy of rPPG measurements, the heart rate derived from rPPG window is compared to those of the reference signals captured concurrently with the videos. We conducted the assessment using two established metrics: Mean Absolute Error (MAE) and Pearson Correlation Coefficient (PCC) of the heart-rate envelope. Additionally, Peak Signal-to-Noise Ratio (PSNR) and Structural Similarity Index (SSIM) were used to evaluate the effectiveness of denoising and inpainting methods in improving image quality.

    \label{sec:evaluationframework}
    
\section{Degradation Module}
    The degradation module simulates a range of spatial factors, visual artifacts, and temporal variations to provide an extensive assessment under varied conditions. Spatial factors include changes in the size of the face crop and distortions due to head or camera movements, requiring stabilization techniques. These factors also account for variations in color bit depth. Visual artifacts simulate common occlusions, such as sunglasses and facemasks, either alone or in combination. Temporal variations mimic fluctuations in video frame rates, reduced frame rates, or random frame dropping, indicative of network or processing lags in remote settings. This module is applied to the facial regions extracted across the entire dataset.

This simulation framework facilitates an in-depth evaluation of the evaluated rPPG methods, offering insights into the operational boundaries and capabilities of the technology across a range of unpredictable environments. 

\subsection{Spatial Factors}
These factors include differences caused by changing the input image size and the impact of alterations in the bit depth of the RGB image.

\paragraph{Facial Region Resolution}

In non-learning rPPG methods, obtaining rPPGs typically revolves around obtaining RGB mean values extracted directly from skin areas of any size. However, employing pre-trained deep learning rPPG methods may impose stringent requirements about input requirements such as square image size or specific resolutions. We conduct experiments at different facial region sizes using simple OpenCV resizing\cite{bradski2000}. As a baseline, we chose 72x72 size due to its balance between computational efficiency and preservation of crucial facial information, as well as compatibility with the selected pre-trained deep learning models.

\begin{figure}[!ht]
    \centering
    \begin{subfigure}{0.23\textwidth}
        \centering
        \includegraphics[width=\linewidth]{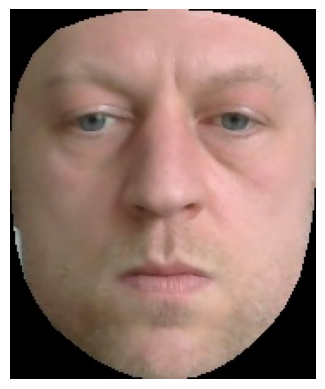}
        \caption{Original Image}
        \label{fig:original_size}
    \end{subfigure}
    \begin{subfigure}{0.23\textwidth}
        \centering
        \includegraphics[width=\linewidth]{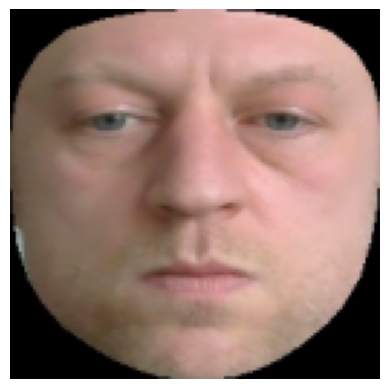}
        \caption{128x128 Image}
        \label{fig:128}
    \end{subfigure}
    \begin{subfigure}{0.23\textwidth}
        \centering
        \includegraphics[width=\linewidth]{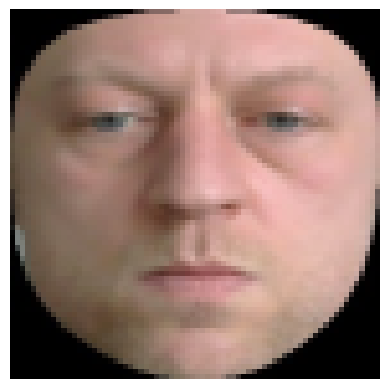}
        \caption{ \B 72x72 Image}
        \label{fig:72}
    \end{subfigure}
    \begin{subfigure}{0.23\textwidth}
        \centering
        \includegraphics[width=\linewidth]{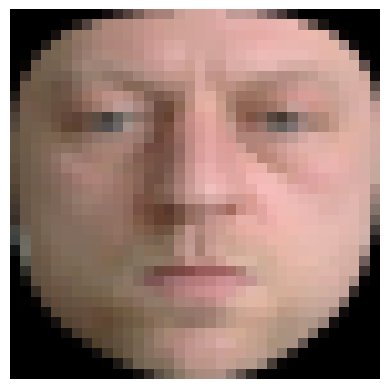}
        \caption{36x36 Image}
        \label{fig:36}
    \end{subfigure}
    \caption{Size variety of the face region}
    \label{fig:size}
    \vspace{-3mm}
\end{figure}

\paragraph{Color Depth Reduction}

Variations in color bit depths in video streams result from a combination of factors, including the diversity of recording equipment, which ranges from smartphones to professional-grade cameras, each offering different levels of color detail. Video compression and encoding standards, such as H.264 and HEVC, further adjust these parameters to balance file size and quality, often reducing both resolution and color bit depth for efficient storage and transmission \cite{mackin2021subjective}. Bandwidth constraints during video streaming need these adjustments to ensure smooth delivery, with the video's resolution and color depth dynamically modified to fit network conditions and device capabilities. Adaptive streaming technologies like MPEG-DASH and Apple's HLS play a role here, optimizing video streams in real-time based on the viewer's internet speed and device performance. Additionally, user preferences for lower data consumption can lead to voluntary reductions in video quality. Such variations can be critical considerations in applications like rPPG, where accurate color information is key, demanding adaptive solutions to maintain signal accuracy across different video qualities. Existing rPPG datasets uniformly employ RGB-8bit color channels, and mostly ignore the potential effects of reducing the color level depth. To simulate this effect we conduct a straightforward mathematical operation to decrease the color depth of a video. Denoting each RGB-8bit frame of the video as $I$ and $nb$ signifies the desired bit number, the operation is defined as follows:

%%  Make this inline and in one line%% [TINO, LE, NHI]
% \begin{equation}
% \begin{aligned}
% I_{nb} &= \left\lfloor \frac{I}{rf} \right\rfloor \cdot rf \\
% rf &= \frac{2^{8}}{2^{nb}} \\
% nb &\in \{6, 4, 2\}
% \end{aligned}
% \end{equation}

\vspace{-5mm}
\begin{center}
\begin{equation}
I_{nb} = \left\lfloor \frac{I}{rf} \right\rfloor \cdot rf, \quad \text{where } rf = \frac{2^{8}}{2^{nb}}, \quad \text{and } nb \in \{6, 4, 2\}
\end{equation}
\end{center}

\begin{figure}[!ht]
    \centering
    \begin{subfigure}{0.23\textwidth}
        \centering
        \includegraphics[width=\linewidth]{Figs/resize/face_resize72.png}
        \caption{8-bit RGB}
        \label{fig:original_rgb}
    \end{subfigure}
    \begin{subfigure}{0.23\textwidth}
        \centering
        \includegraphics[width=\linewidth]{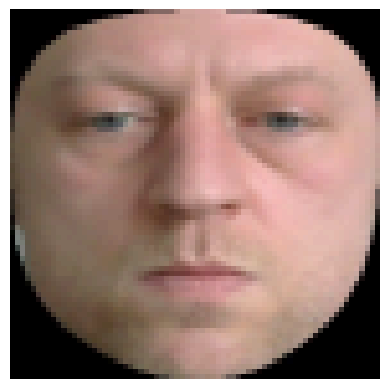}
        \caption{6-bit RGB}
        \label{fig:rgb6}
    \end{subfigure}
    \begin{subfigure}{0.23\textwidth}
        \centering
        \includegraphics[width=\linewidth]{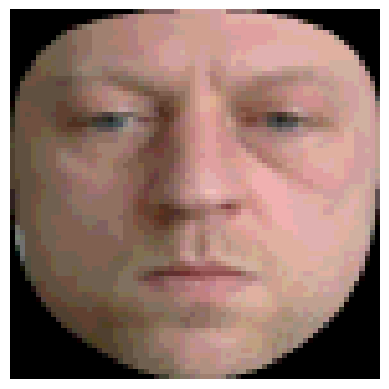}
        \caption{4-bit RGB }
        \label{fig:rgb4}
    \end{subfigure}
    \begin{subfigure}{0.23\textwidth}
        \centering
        \includegraphics[width=\linewidth]{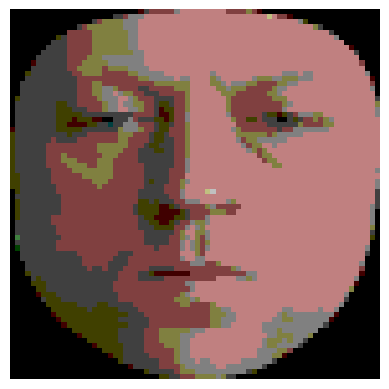}
        \caption{2-bit RGB }
        \label{fig:rgb2}
    \end{subfigure}
    \caption{Color Depth Reduction}
    \label{fig:rgb}
\end{figure}

This process involves determining a reduction factor $rf$ based on the desired bit number and then applying floor division and multiplication to achieve the new color bit frame $I_{nb}$. Noteworthy are the values of $I_{nb}$ still within the range of 0 to 255 similar to $I$. However, the distinction lies in the reduction of intensity levels from the original $2^{8}$ to $2^{nb}$ for each respective channel. In this experiment, $nb$ will take on values of 6, 4, and 2, as visually depicted in Fig~\ref{fig:rgb}.

\paragraph{Facial Image Deterioration}

\begin{figure}[!ht]
    \centering
    \begin{subfigure}{0.23\textwidth}
        \centering
        \includegraphics[width=\linewidth, height=0.18\textheight, keepaspectratio]{Figs/resize/face_resize72.png}
        \caption{72x72 Image}
        \label{fig:original_de}
    \end{subfigure}
    \begin{subfigure}{0.23\textwidth}
        \centering
        \includegraphics[width=\linewidth, height=0.18\textheight, keepaspectratio]{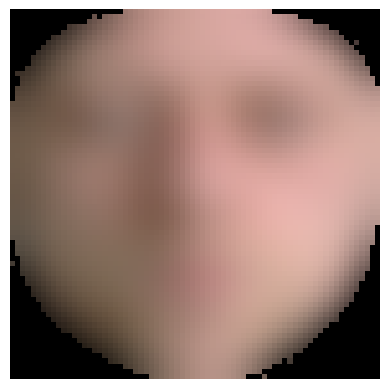}
        \caption{Blur}
        \label{fig:blur}
    \end{subfigure}
    \begin{subfigure}{0.23\textwidth}
        \centering
        \includegraphics[width=\linewidth, height=0.18\textheight, keepaspectratio]{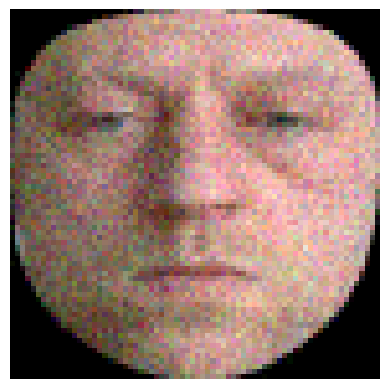}
        \caption{Noise}
        \label{fig:noise}
    \end{subfigure}
    \caption{Facial image deterioration under blur or noise}
    \label{fig:deteriorated}
    \vspace{-3mm}
\end{figure}

Within image deterioration, the two most common types are blur and noise. The generation of these effects involves the utilization of Gaussian functions. Specifically, Gaussian blur is implemented with a square kernel size of 25$\times$25 pixels to induce blur, while Gaussian noise is introduced by applying a distribution with a mean of 0 and a variance of 0.004. We apply blur and noise to the facial regions. Fig~\ref{fig:deteriorated} displays illustrative examples. 

\subsection{Visual Occlusion}

\begin{figure}[!ht]
    \centering
    \begin{subfigure}{0.23\textwidth}
        \centering
        \includegraphics[width=\linewidth]{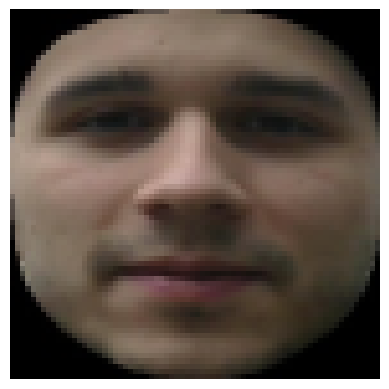}
        \caption{72x72 Image}
        \label{fig:original_vs}
    \end{subfigure}
    \begin{subfigure}{0.23\textwidth}
        \centering
        \includegraphics[width=\linewidth]{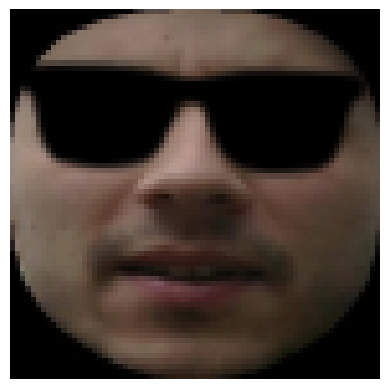}
        \caption{Sunglasses}
        \label{fig:sunglasses}
    \end{subfigure}
    \begin{subfigure}{0.23\textwidth}
        \centering
        \includegraphics[width=\linewidth]{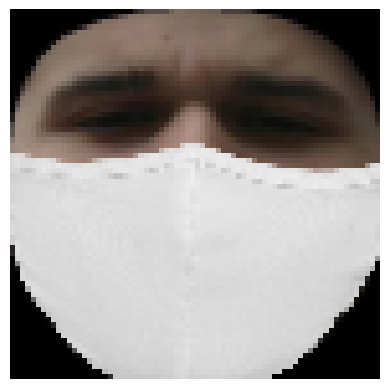}
        \caption{Facemask}
        \label{fig:facemask}
    \end{subfigure}
    \caption{Visual Occlusion Illustration}
    \label{fig:occlusion}
\end{figure}

Integrating heart rate monitoring into real-world scenarios needs to account for occlusions due to everyday items. Leveraging common accessories, such us facemasks, and sunglasses may result in heightened user adoption and sustained monitoring, particularly in outdoor or social settings. Our experiments focus on two commonly worn items specifically, a white facemask and black sunglasses as illustrated in Fig~\ref{fig:occlusion}. The primary objective is to investigate the feasibility of maintaining the original performance of heart rate estimation when these objects cover an individual's face.

The process of integrating sunglasses into facial images involves several steps to ensure a seamless fit and alignment. Initially, the sunglasses undergo resizing to match the dimensions of the target face image. Following this, a translation is performed based on the midpoint of the resized sunglasses and the 6th landmark of the nose bridge, ensuring proper positioning. Subsequently, we align the sunglasses with the eyes, providing a natural and integrated appearance of sunglasses on the facial image. %\cite{manuel2022}

The process of applying facemasks entails a similar procedure. We designed a mask image based on 22 fiducial landmarks of the face delineating the chin and the nose bridge landmark. Subsequently, the same corresponding set of 22 points on the face serves as destination points, outlining the boundary where the mask will be applied. Through the computation of a homography matrix utilizing the destination and source points, the function determines the transformation necessary to accurately map the mask onto the facial landmarks. This transformation, typically involving perspective warping, ensures seamless alignment of the mask with the facial features.

% \begin{comment}
%     \begin{subfigure}{0.24\textwidth}
%         \centering
%         \includegraphics[width=\linewidth]{Figs/occlusion/face_resize72_maskboth_num7.png}
%         \caption{Both}
%         \label{fig:both}
%     \end{subfigure}
% \end{comment}

\subsection{Temporal Variation}

Various devices and video capture systems often function at different frame rates. Also, in real-world applications, data may encounter sporadic sample dropping, attributed to network or computing limitations. In our experiments, we conduct we conducted heart rate computations at different frame rates. Since the original datasets have framerates of either 30 or 25 FPS, we perform experiments at the original rate along with deliberately reduced rates of 20 FPS, 15 FPS, and 10 FPS. We compute the new rates by simply dropping frames at the needed regular rate. In addition, simulated random frame drops into the video streams are introduced to replicate challenges commonly encountered in transmission networks, including package loss, delay, and jitter, which can adversely impact real-time communication applications like video calls. Heart rate calculations were then performed across a spectrum of random frame-drop rates, encompassing dropping 10\%, 20\%, and 50\% of the total video frames. 
    \label{sec:degradation}
    
\section{Mitigation Strategy}
    Although the impact of spatial and temporal deterioration can not be fully overcome, we propose and evaluate different mitigation strategies to improve the those effects with most impact in both spatial (noise and occlusions) and temporal (random frame dropping) artifacts.

\subsection{Noise Mitigation}
In this study, we utilize two denoising methods: Non-local Means (N-LM), a non-learning-based approach~\cite{buades2005}, and NAFNet, a deep learning technique~\cite{liangyu2022}. 

The N-LM algorithm is a denoising technique leveraging natural image statistics to reduce noise, operating under the assumption that similar image patches exhibit comparable noise characteristics. We utilized the implementation provided by OpenCV \textit{(fastNlMeansDenoisingColored)}, with the following parameters: a filter strength of 15 for both the luminance and color components, a template patch size of 7 pixels for weight computation, and a window size of 21 pixels for computing the weighted average for a given pixel. 

\begin{figure}[!ht]
    \centering
    \begin{subfigure}{0.23\textwidth}
        \centering
        \includegraphics[width=\linewidth]{Figs/resize/face_resize72.png}
        \caption{72x72 Image}
        \label{fig:denoise1}
    \end{subfigure}
    \begin{subfigure}{0.23\textwidth}
        \centering
        \includegraphics[width=\linewidth]{Figs/deter/face_resize72_noise.png}
        \caption{Noise}
        \label{fig:denoise2}
    \end{subfigure}
    \begin{subfigure}{0.23\textwidth}
        \centering
        \includegraphics[width=\linewidth]{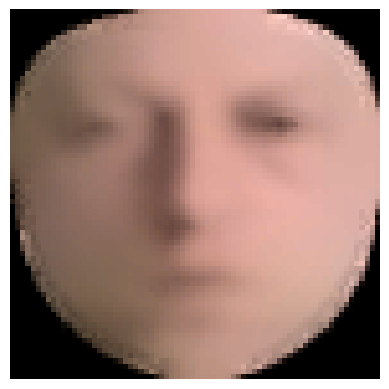}
        \caption{N-LM }
        \label{fig:denoise3}
    \end{subfigure}
    \begin{subfigure}{0.23\textwidth}
        \centering
        \includegraphics[width=\linewidth]{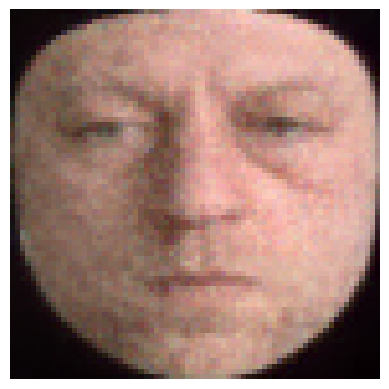}
        \caption{NAFNet}
        \label{fig:denoise4}
    \end{subfigure}
    \caption{Denoising Illustration}
    \label{fig:denoise}
    \vspace{-3mm}
\end{figure}

The NAFNet model is a deep learning network that has demonstrated exceptional performance in tasks like denoising, deblurring, and super-resolution. Developed by customizing a U-net architecture \cite{ronne2015} with skip connections, this pre-trained 32-layer NAFNet on SIDD dataset \cite{abdel2018} was used in our study to assess its effectiveness in denoising datasets used for heart rate estimation. Notably, the key distinction lies in the fact that the denoising process is now exclusively targeted at the face rather than the face with the background. This adjustment allows for a more focused investigation into the impact of noise on skin color, providing valuable insights into the denoising efficacy within this specific context. Figure~\ref{fig:denoise} illustrates an example of denoising utilizing these two methods.

% [TINO COMMENTS] It would be nice to put here the original faces.

\subsection{Visual Occlusion Mitigation}
%Importantly, in both strategies, we do not utilize any specific technique to obtain the mask for the facemask and sunglasses. Instead, we hypothesize that we already possess them to precisely evaluate how effectively both strategies perform without introducing additional variables or biases.

\begin{figure*}[!ht]
  \begin{center}
    \includegraphics[width=\textwidth, height=0.35\textheight, keepaspectratio]{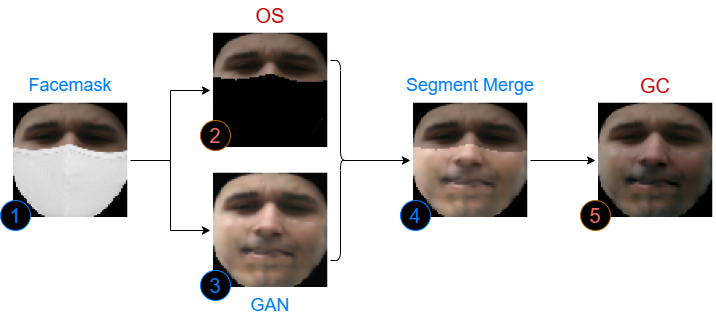}
  \end{center}
  \vspace{-5mm}
  \caption{Illustrations demonstrating facemask visual occlusion mitigation. 1) Facemask image, 2) Original skin image, 3) GAN generated image, 4) Merged image of original and generated skin, 5) Merged image with color transfer from real to generated skin.}
  \label{fig:all_inpaintings}
  \vspace{-3mm}
\end{figure*}

We propose two strategies to mitigate the impact of occlusion on facial regions.
The first method (OS) involves eliminating the facemask and sunglasses replacing them with black color, then only using the skin region of the face as shown in the first step of Fig~\ref{fig:all_inpaintings} for rPPG extraction.

The second method (GC), employs a Generative Adversarial Network (GAN) architecture to generate missing parts of the face and ensure accurate skin color of generative skin region through color transfer. Our GAN comprises a U-Net neural network as the generator, featuring a dual-path design with contracting and expansive paths. The discriminator, modeled after the WGAN-GP architecture, employs a unique loss function and gradient penalty to provide continuous feedback, ensuring detailed and realistic generated facial features. For GAN training, a facial dataset subset from FFHQ was utilized, applying sunglasses and facemasks. The network underwent 100 epochs with a batch size of 16, a fixed learning rate of $2 \times 10^{-4}$ for the generator and $1 \times 10^{-4}$ for the discriminator, and the Adam optimizer for gradient descent optimization.

Following that, the color transfer method is utilized to transfer the color from real skin regions to generated skin regions by leveraging the Lab* color space and the mean and standard deviation of each L*, a*, and b* channel. Figure~\ref{fig:all_inpaintings} illustrates each stage process of the methods applied to facial images in each type of face occlusion.

\subsection{Mitigation of Random Frame Dropping}
\label{strategy:rsm}
\begin{figure*}[!ht]
  \begin{center}
    \includegraphics[width=\textwidth]{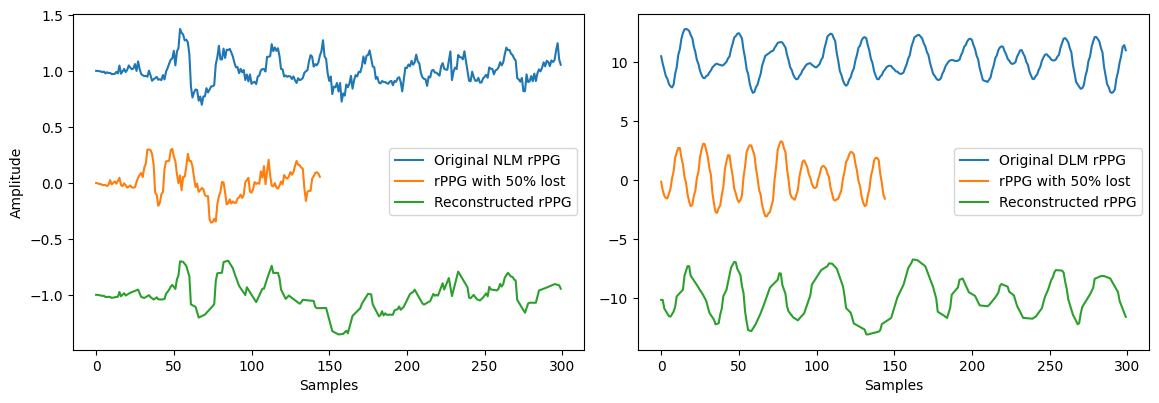}
  \end{center}
  \caption{rPPG Signals in 10 seconds: Non-learning vs Deep Learning with 50\% Frame Drop and Reconstruction (Signals vertically shifted for visual clarity).}
  \label{fig:signal_re}
  \vspace{-3mm}
\end{figure*}

Two distinct mitigation strategies are implemented and evaluated to counteract the impact of these random frame drops. Assuming networked operation, where the videos are transmitted over the network, the first strategy assumes that only the receiver computing device (Rx) is involved. In this case, it recalculates a new frame rate in Frames Per Second (FPS) for the current window based on the number of samples within the window length. In the second strategy, a collaborative approach between the transmitter computing device (Tx) and the receiver is adopted. This involves timestamping each frame and transmitting both the frame and timestamp to the receiver. Consequently, the receiver can reconstruct the signal by either adding the value of the preceding valid frame or adding zeros and refiltering for missing frames. In this research, one-dimensional linear interpolation is employed for monotonically increasing sample points. Fig~\ref{fig:signal_re} depicts 10 seconds rPPG signals before and after randomly dropping 50\% of frames, alongside the reconstructed signal, using both non-learning and deep learning methods.

    \label{sec:mitigation}
    
\section{Results}
    We conduct experiments on both the impact of spatial and temporal degradations and the proposed mitigation experiments in seven different datasets comparing them with baseline results. 

\subsection{Impact of The Spatial Factors}

%\subsubsection{Baseline Selection}  

%%% DISCUSSION

Although our research dataset inputs deviate from the precise training conditions of deep learning methods, we conducted a comparison to assess the adaptability of pre-trained models in handling slightly different input types for the same specific task. It is notable that non-learning methods are expected to compete or even outperform deep learning methods, due to this discrepancy between training and testing data. 

We summarized the experiments in Table~\ref{tab:sizenew}, which shows the performance across a range of different facial region sizes. Based on the results a size of 72$\times$72 was selected for subsequent experiments due to its balanced performance across both non-learning and deep learning methods. This size proved suitable for deployment on EfficientPhys and ContrastPhys, yielding competitive results across diverse datasets while striking a commendable balance between accuracy and resource requirements.

%%% TABLE
\begin{table*}[ht]
\captionsetup{justification = centering}
\caption{Various Image Sizes in Heart Rate Estimation}
\label{tab:sizenew}
\resizebox{\textwidth}{!}{
\begin{tabular}{l *{15}{c}}
\toprule 
\multicolumn{1}{c}{} & \multicolumn{1}{c}{} & \multicolumn{14}{c}{Dataset} \\
\cmidrule{3-16}

\multirow{2}{*}{Size} & \multirow{2}{*}{\begin{tabular}[c]{@{}c@{}}rPPG \\Meth.\end{tabular}}
& \multicolumn{2}{c}{COHFACE}
& \multicolumn{2}{c}{LGI-PPGI} 
& \multicolumn{2}{c}{MAHNOB}
& \multicolumn{2}{c}{PURE}
& \multicolumn{2}{c}{UBFC-rPPG}
& \multicolumn{2}{c}{UCLA-rPPG}
& \multicolumn{2}{c}{UBFC-Phys} \\
\cmidrule(lr){3-4} \cmidrule(lr){5-6} \cmidrule(lr){7-8} \cmidrule(lr){9-10} \cmidrule(lr){11-12} \cmidrule(lr){13-14} 
\cmidrule(lr){15-16}

\multicolumn{1}{c}{} & \multicolumn{1}{c}{}
& \multicolumn{1}{c}{MAE} & \multicolumn{1}{c}{PCC} 
& \multicolumn{1}{c}{MAE} & \multicolumn{1}{c}{PCC}  
& \multicolumn{1}{c}{MAE} & \multicolumn{1}{c}{PCC}  
& \multicolumn{1}{c}{MAE} & \multicolumn{1}{c}{PCC}  
& \multicolumn{1}{c}{MAE} & \multicolumn{1}{c}{PCC} 
& \multicolumn{1}{c}{MAE} & \multicolumn{1}{c}{PCC} 
& \multicolumn{1}{c}{MAE} & \multicolumn{1}{c}{PCC}  \\
\midrule

\multicolumn{1}{c}{} &  \multicolumn{1}{c}{\textcolor{green}{\textbf{NLM}}} & \multicolumn{14}{c}{} \\
\midrule

Org & \multirow{4}{*}{OMIT} & 11.33 & 0.26 & 8.73 & 0.73 & 36.87 & 0.02 & 1.41 & 0.98 & 3.46 & 0.95 & 3.59 & 0.57 & 13.31 & 0.27 \\ 
128 & & 11.28 & 0.26 & 8.76 & 0.73 & 36.94 & 0.03 & 1.41 & 0.98 & 3.45 & 0.95 & 3.59 & 0.57 & 13.30 & 0.27 \\
\B 72 & & \B 11.24 & \B 0.27 & \B 8.79 & \B 0.73 & \B 36.93 & \B 0.02 & \B 1.43 & \B 0.98 & \B 3.49 & \B 0.94 & \B 3.65 & \B 0.56 & \B 13.30 & \B 0.27 \\
36 & & 11.37 & 0.26 & 8.83 & 0.73 & 36.82 & 0.02 & 1.42 & 0.98 & 3.47 & 0.94 & 3.68 & 0.56 & 13.32 & 0.27 \\
\midrule

Org & \multirow{4}{*}{CHROM} & 12.29 & 0.20 & 10.73 & 0.60 & 36.98 & 0.00 & 1.33 & 0.98 & 3.06 & 0.97 & 3.70 & 0.59 & 13.63 & 0.25  \\ 
128 & & 12.36 & 0.20 & 10.85 & 0.60 & 36.95 & -0.01 & 1.35 & 0.98 & 3.06 & 0.97 & 3.69 & 0.59 & 13.63 & 0.25 \\
\B 72 & & \B 12.39 & \B 0.21 & \B 10.87 & \B 0.60 & \B 37.13 & \B -0.02 & \B 1.42 & \B 0.98 & \B 3.05 & \B 0.97 & \B 3.76 & \B 0.59 & \B 13.65 & \B 0.25 \\
36 & & 12.36 & 0.21 & 10.84 & 0.61 & 37.37 & 0.00 & 1.42 & 0.98 & 3.02 & 0.97 & 3.76 & 0.59 & 13.61 & 0.25 \\
\midrule

Org & \multirow{4}{*}{POS} & 11.67 & 0.27 & 6.01 & 0.84 & 36.50 & 0.04 & 1.23 & 0.99 & 2.72 & 0.99 & 3.03 & 0.65 & 13.71 & 0.25 \\ 
128 & & 11.63 & 0.27 & 6.05 & 0.84 & 36.65 & 0.03 & 1.23 & 0.99 & 2.72 & 0.99 & 3.03 & 0.65 & 13.71 & 0.25 \\
\B 72 & & \B 11.58 & \B 0.27 & \B 6.05 & \B 0.84 & \B 36.55 & \B 0.03 & \B 1.25 & \B 0.99 & \B 2.74 & \B 0.99 & \B 3.06 & \B 0.65 & \B 13.69 & \B 0.25 \\
36 & & 11.68 & 0.27 & 6.10 & 0.84 & 36.51 & 0.02 & 1.23 & 0.99 & 2.72 & 0.99 & 3.16 & 0.63 & 13.73 & 0.25 \\
\midrule

\multicolumn{1}{c}{} &  \multicolumn{1}{c}{\textcolor{orange}{\textbf{DLM}}} & \multicolumn{14}{c}{} \\
\midrule

Org & \multirow{4}{*}{\begin{tabular}[c]{@{}c@{}}Phys\\Former\end{tabular}}
& - & - & - & - & - & - & - & - & - & - & - & - & - & - \\ 
128 & & 12.47 & 0.03 & 17.26 & 0.48 & 44.13 & -0.09 & 20.48 & 0.14 & 22.20 & 0.05 & 16.72 & 0.02 & 18.53 & -0.04 \\ 
72 & & - & - & - & - & - & - & - & - & - & - & - & - & - & - \\ 
36 & & - & - & - & - & - & - & - & - & - & - & - & - & - & - \\ 
\midrule

Org & \multirow{4}{*}{\begin{tabular}[c]{@{}c@{}}Efficient\\Phys\end{tabular}}
& - & - & - & - & - & - & - & - & - & - & - & - & - & - \\  
128 & & - & - & - & - & - & - & - & - & - & - & - & - & - & - \\  
\B 72 & & \B 17.46 & \B 0.06 & \B 18.33 & \B 0.37 & \B 38.35 & \B -0.03 & \B 10.12 & \B 0.50 & \B 7.73 & \B 0.69 & \B 7.99 & \B 0.37 & \B 15.93 & \B 0.17 \\
36 & & - & - & - & - & - & - & - & - & - & - & - & - & - & - \\ 
\midrule

Org & \multirow{4}{*}{\begin{tabular}[c]{@{}c@{}}MTTS\\CAN\end{tabular}}
& - & - & - & - & - & - & - & - & - & - & - & - & - & - \\ 
128 & & - & - & - & - & - & - & - & - & - & - & - & - & - & -\\ 
72 & & - & - & - & - & - & - & - & - & - & - & - & - & - & - \\ 
36 & & 38.46 & 0.00 & 55.63 & 0.08 & 53.40 & -0.14 & 89.80 & 0.06 & 71.09 & 0.09 & 100.07 & 0.03 & 83.09 & 0.00\\ 
\midrule

Org & \multirow{4}{*}{\begin{tabular}[c]{@{}c@{}}Contrast\\Phys\end{tabular}}
& - & - & - & - & - & - & - & - & - & - & - & - & - & - \\ 
128 & & 9.65 & 0.21 & 13.39 & 0.52 & 46.51 & 0.07 & 10.91 & 0.62 & 3.14 & 0.97 & 6.43 & 0.44 & 16.29 & 0.20 \\
\B 72 & & \B 9.67 & \B 0.22 & \B 13.86 & \B 0.50 & \B 46.52 & \B 0.06 & \B 11.63 & \B 0.59 & \B 3.72 & \B 0.94 & \B 7.13 & \B 0.40 & \B 16.87 & \B 0.17 \\
36 & & 10.20 & 0.14 & 16.17 & 0.37 & 45.50 & 0.10 & 14.59 & 0.38 & 4.91 & 0.87 & 9.32 & 0.31 & 17.98 & 0.15 \\
\bottomrule
\addlinespace
\end{tabular}
}
\scriptsize{The bolded number serves as the baseline result (BS) for subsequent experiments, where the images are resized to a dimension of 72x72 pixels.}
\end{table*}

While some rPPG methods showed marginal improvements on specific datasets at particular sizes, our experimentation on size variations revealed minimal impact on heart rate estimation in the non-learning method. In the deep learning method, the ContrastPhys model exhibited notable flexibility in handling input size variations, offering greater experimentation opportunities compared to models reliant on fixed dataset sizes. However, it's worth noting that the model still requires uniform input sizes, presenting challenges with original extracted images of varying dimensions. The MTTS-CAN model, designed for pulse and respiration estimation, demonstrated comparatively suboptimal performance in our research, potentially due to distinct dataset processing methods.

Among all the datasets The MAHNOB dataset presents a formidable challenge across various rPPG methods and image sizes. This challenge likely stems from the dataset's inherent complexity, which encompasses diverse emotional expressions, varying lighting conditions, facial movements, and ground truth derived from brain signals.

\subsubsection{Experiments on Color Depth Reduction}

%%% DISCUSSION
We compare different methods using frames with different color depths. Table~\ref{tab:rgb} illustrates a consistent trend observed across most datasets, indicating a general decrease in heart rate estimation as the bit depth decreases. The results consistently favor the use of 8-bit color depth, yielding the most favorable outcomes, while the 2-bit color depth consistently produces the lowest results.

%%% TABLE
\begin{table*}[ht]
\caption{Heart Rate Estimation Errors: Baseline \& RGB bit reduction}
\label{tab:rgb}
\resizebox{\textwidth}{!}{
\begin{tabular}{l *{15}{c}}
\toprule 
\multicolumn{1}{c}{} & \multicolumn{1}{c}{} & \multicolumn{14}{c}{Dataset} \\
\cmidrule{3-16}

\multirow{2}{*}{RGB} & \multirow{2}{*}{\begin{tabular}[c]{@{}c@{}}rPPG \\Meth.\end{tabular}}
& \multicolumn{2}{c}{COHFACE}
& \multicolumn{2}{c}{LGI-PPGI} 
& \multicolumn{2}{c}{MAHNOB}
& \multicolumn{2}{c}{PURE}
& \multicolumn{2}{c}{UBFC-rPPG}
& \multicolumn{2}{c}{UCLA-rPPG}
& \multicolumn{2}{c}{UBFC-Phys} \\
\cmidrule(lr){3-4} \cmidrule(lr){5-6} \cmidrule(lr){7-8} \cmidrule(lr){9-10} \cmidrule(lr){11-12} \cmidrule(lr){13-14} 
\cmidrule(lr){15-16}

\multicolumn{1}{c}{}
& \multicolumn{1}{c}{}
& \multicolumn{1}{c}{MAE} & \multicolumn{1}{c}{PCC} 
& \multicolumn{1}{c}{MAE} & \multicolumn{1}{c}{PCC}  
& \multicolumn{1}{c}{MAE} & \multicolumn{1}{c}{PCC}  
& \multicolumn{1}{c}{MAE} & \multicolumn{1}{c}{PCC}  
& \multicolumn{1}{c}{MAE} & \multicolumn{1}{c}{PCC} 
& \multicolumn{1}{c}{MAE} & \multicolumn{1}{c}{PCC} 
& \multicolumn{1}{c}{MAE} & \multicolumn{1}{c}{PCC}  \\
\midrule

\multicolumn{1}{c}{} &  \multicolumn{1}{c}{\textcolor{green}{\textbf{NLM}}} & \multicolumn{14}{c}{} \\
\midrule
\B 8 bit & \multirow{4}{*}{OMIT} & \B 11.24 & \B 0.27 & \B 8.79 & \B 0.73 & \B 36.93 & \B 0.02 & \B 1.43 & \B 0.98 & \B 3.49 & \B 0.94 & \B 3.65 & \B 0.56 & \B 13.30 & \B 0.27 \\ 
6 bit & & 11.54 & 0.27 & 8.93 & 0.72 & 37.21 & 0.00 & 1.40 & 0.98 & 3.54 & 0.94 & 3.55 & 0.59 & 13.35 & 0.27 \\
4 bit & & 13.15 & 0.23 & 9.34 & 0.71 & 36.33 & 0.08 & 1.88 & 0.95 & 3.51 & 0.95 & 4.26 & 0.54 & 13.59 & 0.25 \\
2 bit & & 19.01 & 0.16 & 13.82 & 0.56 & 37.09 & 0.03 & 6.85 & 0.70 & 16.44 & 0.38 & 9.62 & 0.38 & 15.63 & 0.20 \\
\midrule

\B 8 bit & \multirow{4}{*}{CHROM} & \B 12.39 & \B 0.21 & \B 10.87 & \B 0.60 & \B 37.13 & \B -0.02 & \B 1.42 & \B 0.98 & \B 3.05 & \B 0.97 & \B 3.76 & \B 0.59 & \B 13.65 & \B 0.25  \\ 
6 bit & & 12.42 & 0.22 & 10.85 & 0.61 & 37.42 & -0.04 & 1.44 & 0.98 & 3.07 & 0.97 & 3.79 & 0.59 & 13.59 & 0.25 \\
4 bit & & 14.76 & 0.16 & 11.70 & 0.57 & 36.40 & 0.05 & 2.08 & 0.94 & 3.09 & 0.97 & 4.47 & 0.54 & 13.87 & 0.23 \\
2 bit & & 19.13 & 0.08 & 15.00 & 0.53 & 37.08 & 0.02 & 11.36 & 0.46 & 15.11 & 0.43 & 10.70 & 0.35 & 15.76 & 0.18 \\
\midrule

\B 8 bit & \multirow{4}{*}{POS} & \B 11.58 & \B 0.27 & \B 6.05 & \B 0.84 & \B 36.55 & \B 0.03 & \B 1.25 & \B 0.99 & \B 2.74 & \B 0.99 & \B 3.06 & \B 0.65 & \B 13.69 & \B 0.25 \\ 
6 bit & & 11.81 & 0.29 & 5.97 & 0.84 & 37.00 & 0.01 & 1.23 & 0.99 & 2.78 & 0.98 & 3.09 & 0.65 & 13.70 & 0.25 \\
4 bit & & 14.03 & 0.24 & 6.65 & 0.82 & 36.07 & 0.07 & 2.01 & 0.95 & 2.78 & 0.98 & 3.99 & 0.56 & 13.82 & 0.25 \\
2 bit & & 22.01 & 0.17 & 13.47 & 0.59 & 36.89 & 0.01 & 12.84 & 0.46 & 14.31 & 0.44 & 9.89 & 0.39 & 15.69 & 0.20 \\
\midrule

\multicolumn{1}{c}{} &  \multicolumn{1}{c}{\textcolor{orange}{\textbf{DLM}}} & \multicolumn{14}{c}{} \\
\midrule

\B 8 bit & \multirow{4}{*}{\begin{tabular}[c]{@{}c@{}}Efficient\\Phys\end{tabular}}
& \B 17.46 & \B 0.06 & \B 18.33 & \B 0.37 & \B 38.35 & \B -0.03 & \B 10.12 & \B 0.50 & \B 7.73 & \B 0.69 & \B 7.99 & \B 0.37 & \B 15.93 & \B 0.17 \\ 
6 bit & & 17.50 & 0.05 & 18.46 & 0.38 & 37.96 & -0.03 & 10.95 & 0.45 & 8.63 & 0.65 & 8.52 & 0.34 & 16.26 & 0.16 \\
4 bit & & 17.92 & 0.07 & 20.17 & 0.32 & 37.61 & 0.03 & 16.08 & 0.23 & 14.92 & 0.42 & 12.34 & 0.18 & 18.24 & 0.07 \\
2 bit & & 27.94 & 0.00 & 26.07 & 0.12 & 48.73 & -0.01 & 26.34 & -0.03 & 35.97 & 0.15 & 25.77 & -0.02 & 22.39 & 0.03 \\
\midrule

\B 8 bit & \multirow{4}{*}{\begin{tabular}[c]{@{}c@{}}Contrast\\Phys\end{tabular}}
& \B 9.67 & \B 0.22 & \B 13.86 & \B 0.50 & \B 46.52 & \B 0.06 & \B 11.63 & \B 0.59 & \B 3.72 & \B 0.94 & \B 7.13 & \B 0.40 & \B 16.87 & \B 0.17 \\ 
6 bit & & 9.66 & 0.22 & 13.78 & 0.51 & 46.60 & 0.05 & 11.84 & 0.58 & 3.73 & 0.94 & 7.28 & 0.40 & 16.94 & 0.17 \\
4 bit & & 10.09 & 0.18 & 14.73 & 0.43 & 46.24 & 0.07 & 15.42 & 0.43 & 3.64 & 0.95 & 8.17 & 0.36 & 17.31 & 0.17 \\
2 bit & & 11.28 & 0.05 & 16.34 & 0.39 & 43.89 & 0.05 & 26.37 & 0.22 & 10.05 & 0.63 & 12.43 & 0.27 & 18.80 & 0.13 \\
\bottomrule
\addlinespace
\end{tabular}
}

\scriptsize{The bold number represents the 8-bit RGB image at 72x72 pixels, reflecting the previous baseline result in Table~\ref{tab:sizenew}.}
\vspace{-3mm}
\end{table*}

Remarkably, 6-bit depth yields results very close to those obtained from standard 8-bit images suggesting that there might be room for optimization or compression in image processing pipelines without sacrificing significant accuracy in heart rate estimation. This insight could be particularly valuable in resource-constrained environments, such as mobile or wearable devices, where reducing the bit depth can lead to substantial savings in memory and processing power.

Additionally, it's noteworthy that the EfficientPhys method appears to be particularly sensitive to decreasing bit depth, as evidenced by the more pronounced impact observed in its performance. 

\subsubsection{Experiments on image degradation and mitigation strategies}

We perform experiments for different image degradation and mitigation strategies. 
To address the degradation caused by noise, we applied two denoising methods: NAFNet and N-LM and show a summary of all results in Table \ref{tab:deter}

%%% TABLE
%%LE: We should change the bold format to something else? The behaviour of ContrastPhys is strange, again.
\begin{table*}[ht]
\captionsetup{justification = centering}
\caption{Heart Rate Estimation Errors: Baseline, Deteriorated, and Denoising Datasets }
\label{tab:deter}
\resizebox{\textwidth}{!}{
\begin{tabular}{l *{15}{c}}
\toprule 
\multicolumn{1}{c}{} & \multicolumn{1}{c}{} & \multicolumn{14}{c}{Dataset} \\
\cmidrule{3-16}

\multirow{2}{*}{Trans.}
& \multirow{2}{*}{\begin{tabular}[c]{@{}c@{}}rPPG \\Meth.\end{tabular}}
& \multicolumn{2}{c}{COHFACE}
& \multicolumn{2}{c}{LGI-PPGI} 
& \multicolumn{2}{c}{MAHNOB}
& \multicolumn{2}{c}{PURE}
& \multicolumn{2}{c}{UBFC-rPPG}
& \multicolumn{2}{c}{UCLA-rPPG}
& \multicolumn{2}{c}{UBFC-Phys}    \\
\cmidrule(lr){3-4} \cmidrule(lr){5-6} \cmidrule(lr){7-8} \cmidrule(lr){9-10} \cmidrule(lr){11-12} \cmidrule(lr){13-14} 
\cmidrule(lr){15-16}

\multicolumn{1}{c}{}& \multicolumn{1}{c}{}
& \multicolumn{1}{c}{MAE} & \multicolumn{1}{c}{PCC} 
& \multicolumn{1}{c}{MAE} & \multicolumn{1}{c}{PCC}  
& \multicolumn{1}{c}{MAE} & \multicolumn{1}{c}{PCC}  
& \multicolumn{1}{c}{MAE} & \multicolumn{1}{c}{PCC}  
& \multicolumn{1}{c}{MAE} & \multicolumn{1}{c}{PCC} 
& \multicolumn{1}{c}{MAE} & \multicolumn{1}{c}{PCC} 
& \multicolumn{1}{c}{MAE} & \multicolumn{1}{c}{PCC}  \\
\midrule

\multicolumn{1}{c}{}  & \multicolumn{1}{c}{\textcolor{green}{\textbf{NLM}}} & \multicolumn{14}{c}{} \\
\midrule

BS & \multirow{4}{*}{OMIT} & 11.24 & 0.27 & 8.79 & 0.73 & 36.93 & 0.02 & 1.43 & 0.98 & 3.49 & 0.94 & 3.65 & 0.56 & 13.30 & 0.27 \\
Blur &  & 11.45 & 0.25 & 8.82 & 0.72 & 36.99 & 0.02 & 1.42 & 0.98 & 3.45 & 0.95 & 3.53 & 0.58 & 13.30 & 0.28 \\

\B Noise &  & \B 41.58 & \B 0.10 & \B 17.10 & \B 0.50 & \B 44.83 & \B 0.02 & \B 17.62 & \B 0.35 & \B 13.62 & \B 0.54 & \B 20.16 & \B 0.27 & \B 18.81 & \B 0.19 \\ 
NAFN &  & 38.61 & 0.10 & 18.07 & 0.50 & 46.10 & 0.00 & 18.52 & 0.37 & 15.10 & 0.49 & 21.50 & 0.28 & 19.43 & 0.19 \\
N-LM &  & 42.68 & 0.11 & 18.54 & 0.49 & 45.31 & 0.05 & \U{17.41} & \U{0.36} & 14.87 & 0.51 & 22.26 & 0.24 & 19.65 & 0.20 \\
\midrule

BS & \multirow{4}{*}{CHROM} & 12.39 & 0.21 & 10.87 & 0.60 & 37.13 & -0.02 & 1.42 & 0.98 & 3.05 & 0.97 & 3.76 & 0.59 & 13.65 & 0.25 \\
Blur &  & 12.29 & 0.20 & 10.48 & 0.62 & 37.52 & -0.04 & 1.40 & 0.98 & 3.04 & 0.97 & 3.70 & 0.59 & 13.59 & 0.25 \\

\B Noise &  & \B 48.04 & \B 0.09 & \B 19.90 & \B 0.42 & \B 46.24 & \B 0.02 & \B 19.58 & \B 0.33 & \B 17.12 & \B 0.44 & \B 19.85 & \B 0.28 & \B 19.49 & \B 0.20 \\ 
NAFN &  & 43.20 & 0.07 & 20.76 & 0.38 & 46.60 & 0.02 & 22.33 & 0.31 & 18.68 & 0.40 & 21.67 & 0.27 & 19.61 & 0.19 \\
N-LM &  & 50.58 & 0.10 & 20.96 & 0.40 & 47.43 & 0.02 & 19.98 & 0.33 & 18.70 & 0.42 & 21.35 & 0.26 & 20.12 & 0.19 \\
\midrule

BS & \multirow{4}{*}{POS} & 11.58 & 0.27 & 6.05 & 0.84 & 36.55 & 0.03 & 1.25 & 0.99 & 2.74 & 0.99 & 3.06 & 0.65 & 13.69 & 0.25 \\
Blur &  & 11.82 & 0.25 & 5.87 & 0.85 & 36.50 & 0.03 & 1.23 & 0.99 & 2.76 & 0.98 & 3.08 & 0.64 & 13.67 & 0.25 \\

\B Noise &  & \B 42.75 & \B 0.10 & \B 16.50 & \B 0.54 & \B 44.55 & \B 0.03 & \B 18.30 & \B 0.35 & \B 13.00 & \B 0.53 & \B 21.44 & \B 0.27 & \B 19.67 & \B 0.20 \\ 
NAFN &  & 39.93 & 0.09 & 17.34 & 0.51 & 45.42 & 0.03 & \U{18.14} & \U{0.37} & 14.02 & 0.50 & 21.99 & 0.28 & 20.23 & 0.20 \\
N-LM &  & 43.94 & 0.09 & 17.76 & 0.51 & 44.83 & 0.05 & 18.69 & 0.33 & 14.01 & 0.50 & 23.22 & 0.24 & 20.39 & 0.21 \\
\midrule

\multicolumn{1}{c}{}  & \multicolumn{1}{c}{\textcolor{orange}{\textbf{DLM}}}& \multicolumn{14}{c}{} \\
\midrule
BS & \multirow{4}{*}{\begin{tabular}[c]{@{}c@{}}Efficient\\Phys\end{tabular}} & 17.46 & 0.06 & 18.33 & 0.37 & 38.35 & -0.03 & 10.12 & 0.50 & 7.73 & 0.69 & 7.99 & 0.37 & 15.93 & 0.17 \\ 
Blur &  & 15.86 & 0.14 & 17.64 & 0.42 & 37.66 & 0.02 & 8.89 & 0.55 & 5.16 & 0.85 & 5.59 & 0.51 & 15.59 & 0.18 \\

\B Noise &  & \B 23.10 & \B 0.00 & \B 27.27 & \B 0.08 & \B 40.04 & \B -0.04 & \B 27.07 & \B -0.01 & \B 32.81 & \B -0.02 & \B 26.06 & \B 0.05 & \B 23.60 & \B -0.01\\ 
NAFN &  & 22.35 & -0.01 & \U{25.13} & \U{0.13} & \U{37.71} & \U{-0.03} & \U{22.40} & \U{0.03} & \U{29.90} & \U{0.04} & 19.54 & 0.04 & \U{22.30} & \U{0.01} \\
N-LM &  & \U{20.46} & \U{0.06} & \U{25.44} & \U{0.12} & \U{39.24} & \U{0.03} & \U{23.84} & \U{0.07} & \U{28.79} & \U{0.10} & 20.81 & 0.03 & \U{22.46} & \U{0.02} \\
\midrule

BS & \multirow{4}{*}{\begin{tabular}[c]{@{}c@{}}Contrast\\Phys\end{tabular}} & 9.67 & 0.22 & 13.86 & 0.50 & 46.52 & 0.06 & 11.63 & 0.59 & 3.72 & 0.94 & 7.13 & 0.40 & 16.87 & 0.17 \\
Blur &  & 9.44 & 0.26 & 13.75 & 0.51 & 46.90 & 0.07 & 11.89 & 0.59 & 3.65 & 0.94 & 6.81 & 0.42 & 17.45 & 0.17 \\

\B Noise &  & \B 10.10 & \B 0.10 & \B 14.06 & \B 0.51 & \B 46.58 & \B 0.08 & \B 26.31 & \B 0.16 & \B 6.11 & \B 0.81 & \B 15.11 & \B 0.17 & \B 21.33 & \B 0.07 \\ 
NAFN &  & 10.54 & 0.08 & 15.30 & 0.43 & 45.37 & 0.07 & 26.36 & 0.16 & 7.18 & 0.75 & 16.85 & 0.13 & 22.82 & 0.05 \\
N-LM &  & 10.26 & 0.14 & 14.78 & 0.47 & 46.36 & 0.08 & \U{25.05} & \U{0.21} & 7.02 & 0.75 & 15.10 & 0.15 & 22.36 & 0.06 \\
\bottomrule
\addlinespace
\end{tabular}
}
\scriptsize{{The term BS represents the normal image at 72x72 pixels, the bold represents the deteriorated result needing improvement, while the underline marks enhancements in both metrics compared to the bold.}}
\end{table*}

%%% DISCUSSION
The results underscore the impact of degradation techniques, such as blurring and noise, on the performance of heart rate estimation methods. Notably, blur tends to have a mild impact, and in certain methods and datasets, a blur effect can even slightly enhance performance. Conversely, noise consistently exerts a more pronounced effect on performance.

When considering deteriorated factors in heart rate estimation across datasets like COHFACE, LGI-PPGI, UBFC-rPPG, and UCLA-rPPG, the ContrastPhys method demonstrates lower susceptibility to noise compared to other methods, particularly those in the non-learning category. Although EfficientPhys exhibits relatively lower noise susceptibility in the COHFACE dataset compared to non-learning methods, its performance is significantly affected in other datasets, especially in UBFC-rPPG where it has been pre-trained. Despite these variations, non-learning methods consistently outperform deep learning methods across various degradation types and datasets, except in COHFACE noise datasets, showcasing their robustness. However, deep learning methods demonstrate notable resilience to noise, particularly exemplified by ContrastPhys.

Regarding denoising, while NAFNet and N-LM contribute to improved image quality, the corresponding enhancement in heart rate estimation performance is not as pronounced as shown in Table \ref{tab:deter}. For non-learning methods, the denoising strategies fail to demonstrate significant improvement with N-LM in most datasets, with only NAFNet's denoising method showing slight improvement in COHFACE in terms of the MAE metric. For deep learning methods, in the case of EfficientPhys, both NAFNet and N-LM denoising strategies yield marginal performance improvement, while ContrastPhys does not exhibit any improvement despite the application of denoising techniques.

%%%%%%%%%%%%%
In addition to the impact on heart rate estimation, Fig~\ref{fig:quality_ip} presents an average of all datasets and provides a comparative assessment of image quality metrics, including PSNR and SSIM, where higher values generally indicate superior image quality. Notably, the figure demonstrates that noise and blur deteriorate image quality factors by 36\%-37\%. The N-LM and NAFNet methods respectively demonstrate an average improvement in image quality across both metrics, with SSIM increasing by 20\%-25\% and PSNR by 1.84-2.51.

%%% FIG
\begin{figure*}[!htbp]
  \begin{center}
    \includegraphics[width=\textwidth]{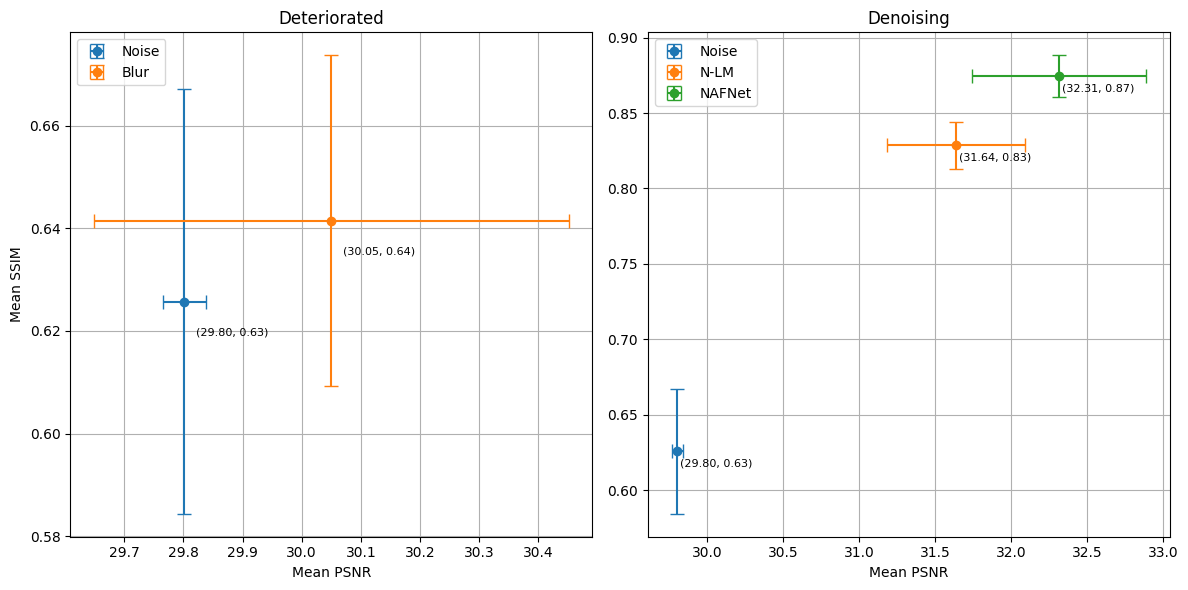}
  \end{center}
  \vspace{-5mm}
  \caption{Average Image Quality Evaluation for Deteriorated \& Mitigation}
  \label{fig:quality_deter}
\end{figure*}

\subsection{Visual Occlusion and Mitigation}

We perform experiments in visual occlusion and mitigation strategies based in inpainting and illustrate them in Table~\ref{tab:ip_all}.

%%% TABLE
%\clearpage
%\onecolumn
%\begin{scriptsize}\tabcolsep=7pt
\begin{table*}[ht]
\captionsetup{justification = centering}
\caption{Heart Rate Estimation Errors: Baseline, Occlusion \& Mitigating Dataset} 
\label{tab:ip_all} 
\resizebox{\textwidth}{!}{
\begin{tabular}{l *{15}{c}}
\toprule 
\multicolumn{1}{c}{} & \multicolumn{1}{c}{} & \multicolumn{14}{c}{Dataset} \\
\cmidrule{3-16}

\multicolumn{1}{c}{}
& \multicolumn{1}{c}{}
& \multicolumn{2}{c}{COHFACE}
& \multicolumn{2}{c}{LGI-PPGI} 
& \multicolumn{2}{c}{MAHNOB}
& \multicolumn{2}{c}{PURE}
& \multicolumn{2}{c}{UBFC-rPPG}
& \multicolumn{2}{c}{UCLA-rPPG}
& \multicolumn{2}{c}{UBFC-Phys}    \\
\cmidrule(lr){3-4} \cmidrule(lr){5-6} \cmidrule(lr){7-8} \cmidrule(lr){9-10} \cmidrule(lr){11-12} \cmidrule(lr){13-14} 
\cmidrule(lr){15-16}

\multicolumn{1}{c}{}& \multicolumn{1}{c}{}
& \multicolumn{1}{c}{MAE} & \multicolumn{1}{c}{PCC} 
& \multicolumn{1}{c}{MAE} & \multicolumn{1}{c}{PCC}  
& \multicolumn{1}{c}{MAE} & \multicolumn{1}{c}{PCC}  
& \multicolumn{1}{c}{MAE} & \multicolumn{1}{c}{PCC}  
& \multicolumn{1}{c}{MAE} & \multicolumn{1}{c}{PCC} 
& \multicolumn{1}{c}{MAE} & \multicolumn{1}{c}{PCC} 
& \multicolumn{1}{c}{MAE} & \multicolumn{1}{c}{PCC}  \\
\midrule

\multicolumn{1}{c}{}  & \multicolumn{1}{c}{\textcolor{green}{\textbf{NL}}} & \multicolumn{14}{c}{} \\
\midrule

BS & \multirow{7}{*}{\begin{tabular}[c]{@{}c@{}}O\\M\\I\\T\end{tabular}} & 11.24 & 0.27 & 8.79 & 0.73 & 36.93 & 0.02 & 1.43 & 0.98 & 3.49 & 0.94 & 3.65 & 0.56 & 13.30 & 0.27 \\*
\cmidrule{3-16}

\B SG &  & \B 11.46 & \B 0.26 & \B 9.45 & \B 0.73 & \B 37.28 & \B 0.03 & \B 1.76 & \B 0.96 & \B 3.64 & \B 0.94 & \B 3.43 & \B 0.57 & \B 13.83 & \B 0.23 \\* 
GC &  & 12.96 & 0.21 & 9.94 & 0.70 & 37.39 & 0.04 & 2.23 & 0.94 & 4.22 & 0.90 & 4.11 & 0.50 & 14.18 & 0.22 \\*
OS &  & 11.96 & 0.23 & 9.95 & 0.70 & 37.58 & 0.02 & 2.17 & 0.94 & 3.83 & 0.92 & 3.44 & 0.57 & 14.00 & 0.22 \\*
\cmidrule{3-16}\B 

\B FM &  & \B 21.89 & \B 0.05 & \B 10.76 & \B 0.69 & \B 36.86 & \B 0.03 & \B 2.99 & \B 0.89 & \B 10.53 & \B 0.48 & \B 7.01 & \B 0.43 & \B 15.47 & \B 0.20 \\* 
GC &  & \U{19.47} & \U{0.07} & \U{9.86} & \U{0.70} & 37.93 & 0.03 & 4.73 & 0.76 & \U{5.84} & \U{0.69} & 8.93 & 0.37 & \U{14.06} & \U{0.26} \\*
OS &  & \U{15.13} & \U{0.10} & \U{8.30} & \U{0.78} & \U{36.77} & \U{0.05} & \U{2.07} & \U{0.95} & \U{4.58} & \U{0.78} & 7.38 & 0.39 & \U{13.72} & \U{0.27} \\*
\midrule

BS & \multirow{7}{*}{\begin{tabular}[c]{@{}c@{}}C\\H\\R\\O\\M\end{tabular}} & 12.39 & 0.21 & 10.87 & 0.60 & 37.13 & -0.02 & 1.42 & 0.98 & 3.05 & 0.97 & 3.76 & 0.59 & 13.65 & 0.25 \\*
\cmidrule{3-16}

\B SG &  & \B 12.67 & \B 0.19 & \B 11.43 & \B 0.61 & \B 37.50 & \B 0.01 & \B 1.75 & \B 0.96 & \B 3.92 & \B 0.89 & \B 3.79 & \B 0.55 & \B 14.04 & \B 0.23 \\* 
GC &  & 13.21 & 0.20 & 11.99 & 0.58 & \U{37.34} & \U{0.03} & 2.20 & 0.94 & \U{3.57} & \U{0.93} & 4.28 & 0.52 & 14.49 & 0.20 \\*
OS &  & 12.68 & 0.21 & 11.39 & 0.61 & 37.72 & 0.00 & 1.85 & 0.96 & \U{3.70} & \U{0.92} & 3.80 & 0.57 & 14.18 & 0.22 \\*
\cmidrule{3-16}

\B FM &  & \B 20.08 & \B 0.06 & \B 11.60 & \B 0.64 & \B 36.91 & \B 0.04 & \B 1.84 & \B 0.96 & \B 11.42 & \B 0.44 & \B 6.87 & \B 0.46 & \B 14.88 & \B 0.25 \\* 
GC &  & 19.32 & 0.06 & \U{10.72} & \U{0.65} & 38.17 & 0.00 & 4.68 & 0.76 & \U{5.22} & \U{0.79} & 7.97 & 0.41 & \U{14.27} & \U{0.26} \\*
OS &  & \U{15.46} & \U{0.11} & \U{9.86} & \U{0.69} & 36.67 & 0.03 & 1.86 & 0.95 & \U{4.11} & \U{0.85} & 6.84 & 0.42 & \U{13.81} & \U{0.27} \\
\midrule
%\pagebreak

BS & \multirow{7}{*}{\begin{tabular}[c]{@{}c@{}}P\\O\\S\end{tabular}} & 11.58 & 0.27 & 6.05 & 0.84 & 36.55 & 0.03 & 1.25 & 0.99 & 2.74 & 0.99 & 3.06 & 0.65 & 13.69 & 0.25 \\*
%\multirow{10}{5pt}{P\\O\\S} & Org & 11.58 & 0.27 & 6.05 & 0.84 & 36.55 & 0.03 & 1.25 & 0.99 & 2.74 & 0.99 & 3.06 & 0.65 & 13.69 & 0.25 \\
\cmidrule{3-16}

\B SG &  & \B 12.10 & \B 0.26 & \B 5.85 & \B 0.86 & \B 36.99 & \B 0.03 & \B 1.69 & \B 0.97 & \B 2.90 & \B 0.97 & \B 3.15 & \B 0.61 & \B 14.00 & \B 0.23 \\* 
GC &  & 13.88 & 0.21 & 6.85 & 0.82 & 36.94 & 0.03 & 1.80 & 0.97 & 3.19 & 0.96 & 3.90 & 0.52 & 14.37 & 0.21 \\*
OS &  & 12.59 & 0.24 & \U{5.81} & \U{0.88} & 37.30 & 0.00 & 1.76 & 0.97 & 3.03 & 0.97 & 3.26 & 0.58 & 14.08 & 0.23 \\*
\cmidrule{3-16}

\B FM &  & \B 17.98 & \B 0.07 & \B 7.74 & \B 0.79 & \B 36.57 & \B 0.02 & \B 2.04 & \B 0.95 & \B 4.72 & \B 0.83 & \B 6.84 & \B 0.38 & \B 14.46 & \B 0.23 \\*
GC &  & 21.58 & 0.09 & 9.20 & 0.73 & 37.79 & 0.04 & 5.26 & 0.70 & \U{3.64} & \U{0.92} & 8.77 & 0.36 & \U{14.18} & \U{0.26} \\*
OS &  & \U{16.04} & \U{0.11} & \U{7.07} & \U{0.81} & 36.60 & 0.03 & 2.00 & 0.95 & \U{2.91} & \U{0.98} & 7.26 & 0.37 & \U{13.71} & \U{0.27} \\
\midrule

\multicolumn{1}{c}{} & \multicolumn{1}{c}{\textcolor{orange}{\textbf{DL}}} & \multicolumn{14}{c}{} \\
\midrule

BS & \multirow{7}{*}{\begin{tabular}[c]{@{}c@{}}E\\P\end{tabular}} & 17.46 & 0.06 & 18.33 & 0.37 & 38.35 & -0.03 & 10.12 & 0.50 & 7.73 & 0.69 & 7.99 & 0.37 & 15.93 & 0.17 \\*
\cmidrule{3-16}

\B SG &  & \B 17.77 & \B 0.06 & \B 19.86 & \B 0.36 & \B 39.10 & \B -0.05 & \B 13.06 & \B 0.36 & \B 14.42 & \B 0.40 & \B 9.81 & \B 0.28 & \B 16.61 & \B 0.16 \\* 
GC &  & 18.56 & 0.04 & 20.25 & 0.35 & \U{38.31} & \U{0.02} & 14.05 & 0.32 & 15.56 & 0.37 & 10.70 & 0.26 & 17.01 & 0.15 \\*
OS &  & 18.31 & 0.04 & \U{19.23} & \U{0.38} & \U{38.82} & \U{0.01} & 14.25 & 0.29 & 14.38 & 0.39 & 10.88 & 0.24 & 16.58 & 0.15 \\*
\cmidrule{3-16}

\B FM &  & \B 21.65 & \B -0.01 & \B 24.40 & \B 0.15 & \B 39.22 & \B 0.00 & \B 22.74 & \B 0.05 & \B 30.68 & \B -0.05 & \B 21.06 & \B 0.02 & \B 22.85 & \B 0.00 \\* 
GC &  & 21.18 & -0.01 & \U{23.09} & \U{0.26} & 38.12 & 0.00 & \U{22.61} & \U{0.08} & \U{29.86} & \U{0.02} & 21.05 & 0.00 & 22.26 & 0.00 \\*
OS &  & \U{39.90} & \U{0.04} & 43.12 & 0.06 & 50.80 & -0.09 & 58.43 & 0.03 & 54.67 & 0.18 & 70.32 & 0.08 & 39.33 & 0.03 \\
\midrule

BS & \multirow{7}{*}{\begin{tabular}[c]{@{}c@{}}C\\P\end{tabular}} &  9.67 & 0.22 & 13.86 & 0.50 & 46.52 & 0.06 & 11.63 & 0.59 & 3.72 & 0.94 & 7.13 & 0.40 & 16.87 & 0.17 \\*
\cmidrule{3-16}

\B SG &  & \B 9.97 & \B 0.16 & \B 16.44 & \B 0.36 & \B 45.66 & \B 0.06 & \B 14.99 & \B 0.44 & \B 3.69 & \B 0.93 & \B 8.41 & \B 0.36 & \B 18.95 & \B 0.13 \\* 
GC &  & 10.31 & 0.16 & \U{14.87} & \U{0.47} & 47.41 & 0.06 & 18.02 & 0.38 & 6.27 & 0.79 & 10.56 & 0.28 & 19.53 & 0.13 \\*
OS &  & 9.97 & 0.16 & 17.05 & 0.32 & \U{46.02} & \U{0.08} & 17.30 & 0.35 & 4.31 & 0.90 & 10.15 & 0.30 & 20.06 & 0.11 \\
\cmidrule{3-16}

\B FM &  & \B 11.56 & \B 0.06 & \B 18.29 & \B 0.28 & \B 46.93 & \B 0.08 & \B 23.17 & \B 0.19 & \B 12.63 & \B 0.36 & \B 16.64 & \B 0.13 & \B 29.10 & \B 0.04 \\* 
GC &  & 11.51 & 0.04 & \U{15.96} & \U{0.41} & 47.11 & 0.09 & 28.51 & 0.16 & \U{9.43} & \U{0.61} & 19.08 & 0.06 & \U{24.47} & \U{0.05} \\*
OS &  & \U{10.82} & \U{0.07} & \U{16.47} & \U{0.42} & \U{46.52} & \U{0.10} & 25.63 & 0.18 & \U{9.89} & \U{0.53} & 19.58 & 0.10 & \U{25.75} & \U{0.07} \\
\bottomrule

\end{tabular}}
\scriptsize{The term "BS" stands as the baseline result for images sized 72x72 pixels without occlusion, the bold represents the result of visual occlusion, while the underline marks enhancements in both metrics compared to the bold. "SG" indicates the presence of sunglasses, while "FM" signifies the presence of facemasks. "OS" refers to the original skin region method, while "GC" represents GAN-OS with the color transfer method.}
\end{table*}
\twocolumn

Our results show that the impact of sunglasses on performance for both learning and non-learning methods was minimal. This could be attributed to the fact that the black color of sunglasses has a lesser effect compared to other colors or eye regions that don't contain important signal information. Hence, mitigation strategies had also minimal impact.

However, the presence of a facemask significantly affects the accuracy of heart rate estimation, with some datasets and methods experiencing a two-fold decrease in accuracy compared to the original. Subsequently, for facemasks, both mitigation strategies showed some mild improvements that varied across datasets and methods.

The original skin region approach exhibited improvement across most datasets and methods and seems to outperform the GC methods. This suggests that while the GC methods generate visually acceptable faces, they may lack certain information crucial for heart rate estimation, resulting in no improvement in accuracy.

However, the two Deep Learning methods demonstrated different trends. While ContrastPhys exhibited a slight improvement in some datasets with both mitigation methods, EfficientPhys performed poorly. This discrepancy may be attributed to the nature of the EfficientPhys model, which relies on temporal information, making the generated skin pixels lack the necessary temporal context. Additionally, the OS method only contains part of the skin region, resulting in a further decrease in performance due to the lack of essential skin pixels.

\begin{figure*}[!thbp]
  \begin{center}
    \includegraphics[width=\textwidth]{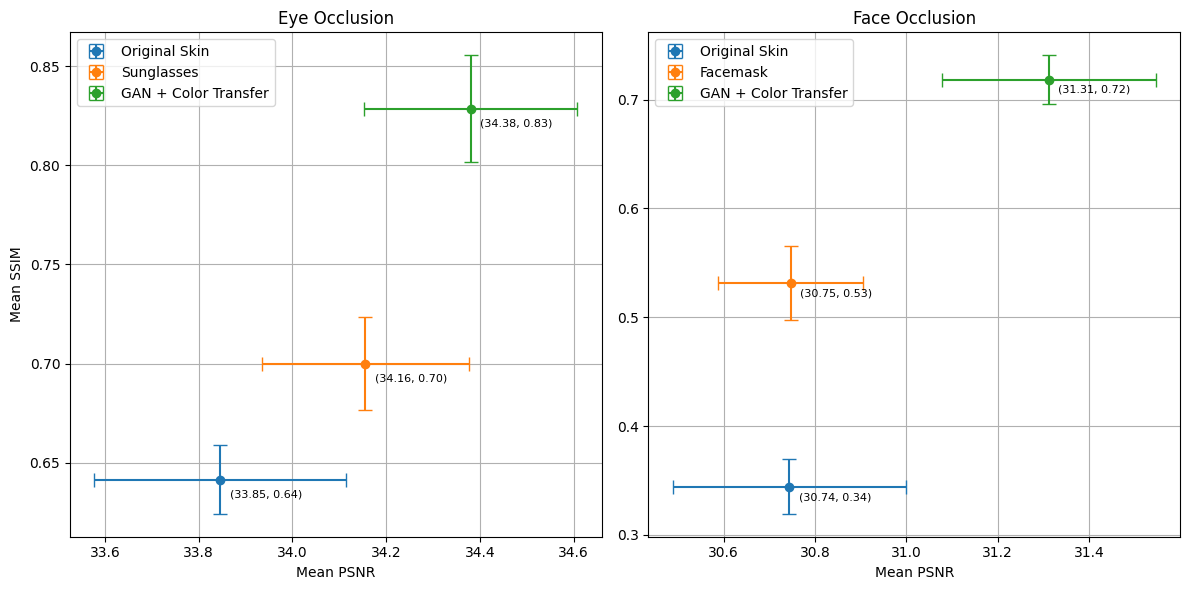}
  \end{center}
  \vspace{-5mm}
  \caption{Average Image Quality Evaluation for Occlusion \& Mitigation}
  \label{fig:quality_ip}
\end{figure*}

%%% DISCUSSION
We compute also results on image quality metrics, and show them in Figure~\ref{fig:quality_ip}, showing an average of all datasets and providing a comparative assessment of image quality metrics. Notably, the figure demonstrates that sunglasses impact image quality by 26\%-33\%, while the facemask has a more significant impact, ranging from 41\%-51\%.

As expected, the first mitigation strategy (OS), which replaces occluded parts with a black background, consistently decreases image quality in all occlusion scenarios. Conversely, the second mitigation strategy (GC) proves to be the most effective in mitigating the impact of occlusions on image quality. Specifically, in scenarios where the face is occluded by a facemask region, the GC strategy improves SSIM by about 19\% and PSNR by approximately 0.56. Similarly, when occluded by sunglasses, the GC strategy improves SSIM by about 13\% and PSNR by approximately 0.22. These results show that image quality is important for visualization purposes, but might not necessarily result in better estimation of rPPG signals.

%%% FIG
%%% DISCUSSION

\subsection{Temporal Variation Analysis}

We conduct experiments in both changes in sample rate and random frame dropping. 
We show experiments for different frame rates in Table \ref{tab:sample_rate}. In the experiments, the COHFACE dataset has an original frame rate of 20 FPS, resulting in one missing result compared to other datasets.

%%% TABLE
\begin{table*}[ht]
\caption{Heart Rate Errors: Sample Rate Variations}
\label{tab:sample_rate}
\resizebox{\textwidth}{!}{
\begin{tabular}{l *{15}{c}}
\toprule 
\multicolumn{1}{c}{} & \multicolumn{1}{c}{} & \multicolumn{14}{c}{Dataset} \\
\cmidrule{3-16}

\multirow{2}{*}{$F_{s}$}
& \multirow{2}{*}{\begin{tabular}[c]{@{}c@{}}rPPG \\Meth.\end{tabular}}
& \multicolumn{2}{c}{COHFACE}
& \multicolumn{2}{c}{LGI-PPGI} 
& \multicolumn{2}{c}{MAHNOB}
& \multicolumn{2}{c}{PURE}
& \multicolumn{2}{c}{UBFC-rPPG}
& \multicolumn{2}{c}{UCLA-rPPG}
& \multicolumn{2}{c}{UBFC-Phys}    \\
\cmidrule(lr){3-4} \cmidrule(lr){5-6} \cmidrule(lr){7-8} \cmidrule(lr){9-10} \cmidrule(lr){11-12} \cmidrule(lr){13-14} 
\cmidrule(lr){15-16}

\multicolumn{1}{c}{}& \multicolumn{1}{c}{}
& \multicolumn{1}{c}{MAE} & \multicolumn{1}{c}{PCC} 
& \multicolumn{1}{c}{MAE} & \multicolumn{1}{c}{PCC}  
& \multicolumn{1}{c}{MAE} & \multicolumn{1}{c}{PCC}  
& \multicolumn{1}{c}{MAE} & \multicolumn{1}{c}{PCC}  
& \multicolumn{1}{c}{MAE} & \multicolumn{1}{c}{PCC} 
& \multicolumn{1}{c}{MAE} & \multicolumn{1}{c}{PCC} 
& \multicolumn{1}{c}{MAE} & \multicolumn{1}{c}{PCC}  \\
\midrule

\multicolumn{1}{c}{} &  \multicolumn{1}{c}{\textcolor{green}{\textbf{NLM}}} & \multicolumn{14}{c}{} \\
\midrule

\B BS & \B \multirow{4}{*}{OMIT} & \B \multirow{2}{*}{11.24} & \B \multirow{2}{*}{0.27} & \B 8.79 & \B 0.73 & \B 36.93 & \B 0.02 & \B 1.43 & \B 0.98 & \B 3.49 & \B 0.94 & \B 3.65 & \B 0.56 & \B 13.30 & \B 0.27 \\ 
20 & &  &  & 11.99 & 0.58 & 37.34 & 0.01 & 1.65 & 0.95 & 4.10 & 0.89 & 4.12 & 0.55 & 13.05 & 0.27 \\
15 & & 12.57 & 0.21 & 11.87 & 0.63 & 37.36 & 0.02 & 1.22 & 0.98 & 3.48 & 0.94 & 4.07 & 0.54 & 13.13 & 0.27 \\
10 & & 13.26 & 0.24 & 11.50 & 0.64 & 37.83 & 0.00 & 1.26 & 0.98 & \U{3.35} & \U{0.95} & 4.40 & 0.49 & 13.17 & 0.27 \\

\midrule
\B BS & \B \multirow{4}{*}{CHROM} & \B \multirow{2}{*}{12.39} & \B \multirow{2}{*}{0.21} & \B \B 10.87 & \B \B 0.60 & \B \B 37.13 & \B \B -0.02 & \B \B 1.42 & \B \B 0.98 & \B \B 3.05 & \B \B 0.97 & \B \B 3.76 & \B \B 0.59 & \B \B 13.65 & \B \B 0.25  \\ 
20 & &  &  & 12.19 & 0.57 & 37.50 & -0.01 & 1.67 & 0.95 & 3.81 & 0.91 & 4.22 & 0.56 & \U{13.31} & \U{0.26} \\
15 & & 13.93 & 0.13 & 13.24 & 0.59 & 37.62 & 0.00 & 1.24 & 0.98 & 3.12 & 0.96 & 4.16 & 0.57 & \U{13.40} & \U{0.26} \\
10 & & 13.96 & 0.21 & 12.87 & 0.57 & 38.37 & -0.01 & \U{1.18} & \U{0.99} & 3.05 & 0.97 & 4.95 & 0.48 & 13.46 & 0.25 \\
\midrule

\B BS & \B \multirow{4}{*}{POS} & \B \multirow{2}{*}{11.58} & \B \multirow{2}{*}{0.27} & \B \B 6.05 & \B \B 0.84 & \B \B 36.55 & \B \B 0.03 & \B \B 1.25 & \B \B 0.99 & \B \B 2.74 & \B \B 0.99 & \B \B 3.06 & \B \B 0.65 & \B \B 13.69 & \B \B 0.25 \\ 
20 & &  &  & 6.93 & 0.79 & 37.63 & -0.01 & 1.39 & 0.98 & 3.08 & 0.96 & 4.02 & 0.54 & \U{13.31} & \U{0.26} \\
15 & & 13.38 & 0.20 & 7.85 & 0.76 & 37.40 & 0.02 & 1.12 & 0.99 & 2.79 & 0.98 & 4.01 & 0.53 & 13.43 & 0.25 \\
10 & & 14.65 & 0.23 & 7.53 & 0.76 & 38.42 & -0.01 & 1.29 & 0.98 & 2.69 & 0.99 & 4.46 & 0.48 & \U{13.57} & \U{0.26} \\
\midrule

\multicolumn{1}{c}{} & \multicolumn{1}{c}{\textcolor{orange}{\textbf{DLM}}} & \multicolumn{14}{c}{} \\
\midrule

\B BS & \multirow{4}{*}{\begin{tabular}[c]{@{}c@{}}Efficient\\Phys\end{tabular}} 
& \B \multirow{2}{*}{17.46} & \B \multirow{2}{*}{0.06} & \B 18.33 & \B 0.37 & \B 38.35 & \B -0.03 & \B 10.12 & \B 0.50 & \B 7.73 & \B 0.69 & \B 7.99 & \B 0.37 & \B 15.93 & \B 0.17 \\ 
20 & &  &  & \U{18.11} & \U{0.40} & \U{37.43} & \U{-0.01} & 9.62 & 0.45 & 9.33 & 0.59 & \U{6.16} & \U{0.48} & \U{15.07} & \U{0.18} \\
15 & & \U{15.29} & \U{0.07} & \U{17.03} & \U{0.41} & 36.85 & -0.03 & 8.81 & 0.49 & \U{6.55} & \U{0.74} & \U{4.97} & \U{0.59} & \U{14.51} & \U{0.20} \\
10 & & \U{11.93} & \U{0.15} & 20.13 & 0.25 & \U{36.82} & \U{0.05} & 9.42 & 0.39 & 13.27 & 0.38 & \U{5.62} & \U{0.52} & 14.74 & 0.16 \\
\midrule

\B BS & \multirow{4}{*}{\begin{tabular}[c]{@{}c@{}}Contrast\\Phys\end{tabular}}
& \B \multirow{2}{*}{9.67} & \B \multirow{2}{*}{0.22} & \B 13.86 & \B 0.50 & \B 46.52 & \B 0.06 & \B 11.63 & \B 0.59 & \B 3.72 & \B 0.94 & \B 7.13 & \B 0.40 & \B 16.87 & \B 0.17 \\ 
20 & &  &  & 17.72 & 0.24 & 34.92 & 0.06 & 8.46 & 0.49 & 21.55 & -0.11 & \U{5.78} & \U{0.50} & 13.68 & 0.17 \\
15 & & 15.21 & -0.02 & 24.50 & -0.05 & 45.43 & 0.01 & 10.30 & 0.20 & 38.45 & -0.35 & 12.94 & -0.09 & 22.86 & -0.12 \\
10 & & 23.03 & -0.04 & 33.37 & -0.09 & 52.95 & 0.03 & 20.26 & -0.05 & 48.38 & -0.17 & 25.03 & -0.11 & 31.01 & -0.01 \\
\bottomrule
\addlinespace
\end{tabular}
}

\scriptsize{The bold denotes the baseline result (BS) for images sized 72x72 pixels with the original sample rate while the underlined mark unexpectedly improves in both metrics compared to the BS.}
\end{table*}

When simulating downsampled frame rates from the original down to 10 FPS, the results for NLM methods remain relatively consistent, with minor fluctuations across datasets. There are some unexpected improvements, but they are not substantial. However, in DLM methods, EfficientPhys exhibits unusual behavior, demonstrating better performance at lower frame rates in some datasets, even in its pre-trained UBFC-rPPG dataset with 30 FPS. This anomaly may be due to the lower frame rate providing more information on the differences between the two frames, which benefits the model, but requires a more in-depth investigation.

On the other hand, ContrastPhys showcases a decreasing trend in results corresponding to the frame rate, especially on its pre-trained dataset UBFC-rPPG, where it shows a clear decreasing trend. This behavior suggests that ContrastPhys relies heavily on power spectrum densities of rPPG signals, and with too few samples, this method does not perform well.

% -----------------------------------------------------------------------------------------------------------------------------
%%% DISCUSSION
Additionally we summarize results from random frame dropping experiments in Table \ref{tab:rsm}-\ref{tab:rsm2} across with both mitigation strategies. 

%%% TABLE
\begin{table*}[ht]
\caption{Heart Rate Errors: Missing Randomized Samples Pt. 1}
\label{tab:rsm}
\resizebox{\textwidth}{!}{
\begin{tabular}{c *{19}{c}}
\toprule 
\multicolumn{1}{c}{} & \multicolumn{1}{c}{} & \multicolumn{18}{c}{Dataset} \\
\cmidrule{3-20}

\multirow{3}{*}{\begin{tabular}[c]{@{}c@{}}Lost \\(\%)\end{tabular}}
& \multirow{3}{*}{\begin{tabular}[c]{@{}c@{}}rPPG \\Meth.\end{tabular}}
& \multicolumn{6}{c}{COHFACE}
& \multicolumn{6}{c}{LGI-PPGI} 
& \multicolumn{6}{c}{MAHNOB} \\
\cmidrule(lr){3-8} \cmidrule(lr){9-14} \cmidrule(lr){15-20}

& 
& \multicolumn{2}{c}{Strategy 0} & \multicolumn{2}{c}{Strategy 1} & \multicolumn{2}{c}{Strategy 2}
& \multicolumn{2}{c}{Strategy 0} & \multicolumn{2}{c}{Strategy 1} & \multicolumn{2}{c}{Strategy 2}
& \multicolumn{2}{c}{Strategy 0} & \multicolumn{2}{c}{Strategy 1} & \multicolumn{2}{c}{Strategy 2}
\\
\cmidrule(lr){3-4} \cmidrule(lr){5-6} \cmidrule(lr){7-8} \cmidrule(lr){9-10} \cmidrule(lr){11-12} \cmidrule(lr){13-14} 
\cmidrule(lr){15-16} \cmidrule(lr){17-18} \cmidrule(lr){19-20}

& 
& \multicolumn{1}{c}{MAE} & \multicolumn{1}{c}{PCC} 
& \multicolumn{1}{c}{MAE} & \multicolumn{1}{c}{PCC} 
& \multicolumn{1}{c}{MAE} & \multicolumn{1}{c}{PCC} 
& \multicolumn{1}{c}{MAE} & \multicolumn{1}{c}{PCC} 
& \multicolumn{1}{c}{MAE} & \multicolumn{1}{c}{PCC} 
& \multicolumn{1}{c}{MAE} & \multicolumn{1}{c}{PCC} 
& \multicolumn{1}{c}{MAE} & \multicolumn{1}{c}{PCC} 
& \multicolumn{1}{c}{MAE} & \multicolumn{1}{c}{PCC} 
& \multicolumn{1}{c}{MAE} & \multicolumn{1}{c}{PCC} 
\\
\midrule

\multicolumn{1}{c}{} &  \multicolumn{1}{c}{\textcolor{green}{\textbf{NLM}}}& \multicolumn{18}{c}{} \\
\midrule

%0 & \multirow{4}{*}{OMIT} 
\B 0 & \multirow{4}{*}{\begin{tabular}[c]{@{}c@{}}O\\M\\I\\T\end{tabular}} 
& \B 11.24 & \B 0.27 &  &  &  &  & \B 8.79 & \B 0.97 &  &  &  &  & \B 36.93 & \B 0.02 &  &  &  &  \\
10 & & 13.45 & 0.19 & 12.64 & 0.25 & 11.52 & 0.26 & 14.69 & 0.67 & 10.02 & 0.70 & 9.52 & 0.69 & 36.94 & 0.02 & 37.02 & 0.03 & 36.58 & 0.04 \\
20 & & 16.87 & 0.05 & 13.49 & 0.22 & 11.75 & 0.26 & 21.93 & 0.58 & 12.21 & 0.63 & 9.61 & 0.69 & 37.16 & 0.00 & 37.39 & 0.01 & 36.74 & 0.03 \\
50 & & 24.33 & -0.04 & 26.69 & 0.13 & 12.18 & 0.25 & 46.72 & 0.00 & 16.36 & 0.61 & 11.54 & 0.63 & 38.63 & -0.02 & 38.10 & 0.00 & 36.61 & 0.05 \\
\midrule

%0 & \multirow{4}{*}{CHROM} 
\B 0 & \multirow{4}{*}{\begin{tabular}[c]{@{}c@{}}C\\H\\R\\O\end{tabular}} 
& \B 12.39 & \B 0.21 &  &  &  &  & \B 10.87 & \B 0.60 &  &  &  &  & \B 37.13 & \B -0.02 &  &  &  &  \\
10 & & 14.92 & 0.11 & 13.47 & 0.19 & 12.87 & 0.21 & 16.51 & 0.57 & 11.83 & 0.60 & 11.10 & 0.59 & 36.65 & 0.02 & 37.18 & 0.00 & 37.08 & -0.02 \\
20 & & 18.02 & 0.00 & 14.77 & 0.16 & 13.16 & 0.21 & 22.82 & 0.51 & 13.78 & 0.55 & 11.49 & 0.58 & 37.26 & -0.01 & 37.71 & -0.01 & 37.01 & -0.01 \\
50 & & 22.91 & -0.02 & 30.99 & 0.11 & 13.19 & 0.18 & 45.95 & 0.00 & 17.58 & 0.57 & 12.78 & 0.55 & 39.18 & -0.03 & 38.54 & 0.00 & 36.93 & 0.00 \\
\midrule

%0 & \multirow{4}{*}{POS} 
\B 0 & \multirow{4}{*}{\begin{tabular}[c]{@{}c@{}}P\\O\\S\end{tabular}} 
& \B 11.58 & \B 0.27 &  &  &  &  & \B 6.05 & \B 0.84 &  &  &  &  & \B 36.55 & \B 0.03 &  &  &  &  \\
10 & & 13.59 & 0.23 & 12.95 & 0.25 & 11.88 & 0.27 & 12.44 & 0.82 & 7.45 & 0.80 & 6.64 & 0.81 & 36.69 & 0.02 & 37.01 & 0.01 & 36.43 & 0.04 \\
20 & & 17.28 & 0.13 & 14.04 & 0.22 & 12.01 & 0.27 & 20.99 & 0.76 & 9.37 & 0.74 & 6.76 & 0.81 & 36.66 & -0.02 & 37.25 & 0.01 & 36.46 & 0.05 \\
50 & & 27.03 & -0.03 & 27.61 & 0.12 & 12.33 & 0.25 & 49.26 & -0.05 & 15.74 & 0.56 & 7.86 & 0.76 & 37.82 & 0.01 & 38.50 & -0.01 & 36.65 & 0.04 \\
\midrule

\multicolumn{1}{c}{} &  \multicolumn{1}{c}{\textcolor{orange}{\textbf{DLM}}}& \multicolumn{18}{c}{} \\
\midrule

%0 & \multirow{4}{*}{\begin{tabular}[c]{@{}c@{}}Efficient\\Phys\end{tabular}} 
\B 0 & \multirow{4}{*}{\begin{tabular}[c]{@{}c@{}}E\\P\end{tabular}} 
%0 & \multirow{4}{*}{EP}
& \B 17.46 & \B 0.06 &  &  &  &  & \B 18.33 & \B 0.37 &  &  &  &  & \B 38.35 & \B -0.03 &  &  &  &  \\
10 & & 17.72 & 0.06 & 16.35 & 0.08 & 15.19 & 0.07 & 19.89 & 0.38 & 18.09 & 0.43 & 17.71 & 0.43 & 38.49 & -0.03 & 38.48 & -0.06 & 38.09 & -0.03 \\
20 & & 19.84 & 0.03 & 15.61 & 0.11 & 13.97 & 0.10 & 21.49 & 0.38 & 17.49 & 0.47 & 16.69 & 0.46 & 38.19 & -0.02 & 37.36 & 0.03 & 37.08 & 0.04 \\
50 & & 25.69 & 0.01 & 13.65 & 0.07 & 12.71 & 0.12 & 36.31 & -0.11 & 21.37 & 0.30 & 20.91 & 0.25 & 38.20 & 0.01 & 36.97 & 0.01 & 37.98 & 0.06 \\
\midrule

%0 & \multirow{4}{*}{\begin{tabular}[c]{@{}c@{}}Contrast\\Phys\end{tabular}} 
\B 0 & \multirow{4}{*}{\begin{tabular}[c]{@{}c@{}}C\\P\end{tabular}} 
%0 & \multirow{4}{*}{CP}  
& \B 9.67 & \B 0.22 &  &  &  &  & \B 13.86 & \B 0.50 &  &  &  &  & \B 46.52 & \B 0.06 &  &  &  &  \\
10 & & 10.76 & 0.09 & 11.62 & 0.15 & 11.03 & 0.14 & 17.65 & 0.33 & 16.83 & 0.36 & 16.10 & 0.36 & 46.00 & 0.08 & 43.64 & 0.07 & 43.14 & 0.08 \\
20 & & 13.14 & -0.10 & 14.33 & 0.03 & 13.44 & 0.03 & 21.14 & 0.18 & 18.76 & 0.30 & 18.27 & 0.21 & 46.42 & 0.07 & 41.42 & 0.06 & 41.15 & 0.05 \\
50 & & 12.25 & 0.01 & 30.62 & 0.06 & 18.86 & 0.02 & 29.28 & -0.41 & 30.48 & 0.02 & 27.12 & 0.00 & 43.53 & 0.10 & 41.75 & 0.06 & 39.57 & 0.08 \\
\bottomrule
\addlinespace
\end{tabular}
}
\scriptsize{The bold denotes the baseline result (BS) for images sized 72x72 pixels with 0\% lost sample.}
\end{table*}

%%% TABLE

%\onecolumn
\begin{table*}[ht]
\caption{Heart Rate Errors: Missing Randomized Samples Pt. 2} 
\label{tab:rsm2} 
\resizebox*{!}{0.85\textheight}{
\begin{tabular}{c *{13}{c}}
\toprule
\multicolumn{1}{c}{} & \multicolumn{1}{c}{} & \multicolumn{12}{c}{Dataset} \\*
\cmidrule{3-14}

\multirow{3}{*}{\begin{tabular}[c]{@{}c@{}}Lost \\(\%)\end{tabular}}
& \multirow{3}{*}{\begin{tabular}[c]{@{}c@{}}rPPG \\Meth.\end{tabular}}
& \multicolumn{6}{c}{PURE}
& \multicolumn{6}{c}{UBFC-rPPG} \\*
\cmidrule(lr){3-8} \cmidrule(lr){9-14}

& 
& \multicolumn{2}{c}{Strategy 0} & \multicolumn{2}{c}{Strategy 1} & \multicolumn{2}{c}{Strategy 2}
& \multicolumn{2}{c}{Strategy 0} & \multicolumn{2}{c}{Strategy 1} & \multicolumn{2}{c}{Strategy 2}\\*
\cmidrule(lr){3-4} \cmidrule(lr){5-6} \cmidrule(lr){7-8} \cmidrule(lr){9-10} \cmidrule(lr){11-12} \cmidrule(lr){13-14} 

& 
& \multicolumn{1}{c}{MAE} & \multicolumn{1}{c}{PCC} 
& \multicolumn{1}{c}{MAE} & \multicolumn{1}{c}{PCC} 
& \multicolumn{1}{c}{MAE} & \multicolumn{1}{c}{PCC} 
& \multicolumn{1}{c}{MAE} & \multicolumn{1}{c}{PCC} 
& \multicolumn{1}{c}{MAE} & \multicolumn{1}{c}{PCC} 
& \multicolumn{1}{c}{MAE} & \multicolumn{1}{c}{PCC} \\
\midrule

\multicolumn{1}{c}{} &  \multicolumn{1}{c}{\textcolor{green}{\textbf{NLM}}}& \multicolumn{12}{c}{} \\
\midrule

\B 0 & \multirow{4}{*}{OMIT} 
& \B 1.43 & \B 0.98 &  &  &  &  & \B 3.49 & \B 0.94 &  &  &  &  \\*
10 & & 7.32 & 0.96 & 2.59 & 0.97 & 1.46 & 0.97 & 8.80 & 0.90 & 5.38 & 0.94 & 3.46 & 0.94 \\*
20 & & 16.57 & 0.93 & 3.18 & 0.96 & 1.43 & 0.97 & 20.92 & 0.82 & 6.32 & 0.93 & 3.48 & 0.94 \\*
50 & & 55.57 & 0.31 & 7.45 & 0.85 & 1.69 & 0.96 & 62.75 & 0.15 & 12.83 & 0.76 & 3.65 & 0.94 \\
\midrule

\B 0 & \multirow{4}{*}{CHROM} 
& \B 1.42 & \B 0.98 &  &  &  &  & \B 3.05 & \B 0.97 &  &  &  &  \\*
10 & & 7.43 & 0.95 & 2.61 & 0.97 & 1.40 & 0.98 & 8.37 & 0.94 & 4.96 & 0.96 & 3.06 & 0.97 \\*
20 & & 16.61 & 0.93 & 3.27 & 0.96 & 1.48 & 0.97 & 20.66 & 0.87 & 6.04 & 0.94 & 3.18 & 0.96 \\*
50 & & 55.93 & 0.37 & 7.17 & 0.87 & 1.63 & 0.96 & 62.34 & 0.06 & 13.54 & 0.65 & 3.29 & 0.96 \\
\midrule

\B 0 & \multirow{4}{*}{POS} 
& \B 1.25 & \B 0.99 &  &  &  &  & \B 2.74 & \B 0.99 &  &  &  &   \\*
10 & & 7.22 & 0.98 & 2.45 & 0.98 & 1.26 & 0.99 & 8.16 & 0.96 & 4.78 & 0.97 & 2.78 & 0.98 \\*
20 & & 16.78 & 0.94 & 3.28 & 0.96 & 1.23 & 0.99 & 20.27 & 0.92 & 5.95 & 0.95 & 2.80 & 0.98 \\*
50 & & 56.02 & 0.31 & 8.53 & 0.74 & 1.64 & 0.97 & 65.89 & 0.20 & 12.19 & 0.78 & 3.03 & 0.97 \\
\midrule

\multicolumn{1}{c}{} &  \multicolumn{1}{c}{\textcolor{orange}{\textbf{DLM}}}& \multicolumn{12}{c}{} \\
\midrule

\B 0 & \multirow{4}{*}{\begin{tabular}[c]{@{}c@{}}Efficient\\Phys\end{tabular}} 
& \B 10.12 & \B 0.50 &  &  &  &  & \B 7.73 & \B 0.69 &  &  &  &  \\*
10 & & 14.72 & 0.32 & 12.23 & 0.33 & 10.31 & 0.45 & 13.86 & 0.57 & 9.90 & 0.68 & 7.71 & 0.71 \\*
20 & & 19.45 & 0.32 & 11.01 & 0.47 & 9.06 & 0.54 & 24.00 & 0.43 & 11.69 & 0.62 & 8.58 & 0.65 \\*
50 & & 40.61 & -0.10 & 14.27 & 0.32 & 10.93 & 0.30 & 44.26 & -0.17 & 19.49 & 0.45 & 14.46 & 0.43 \\
\midrule
 
\B 0 & \multirow{4}{*}{\begin{tabular}[c]{@{}c@{}}Contrast\\Phys\end{tabular}} 
&  \B 11.63 & \B 0.59 &  &  &  &  & \B 3.72 & \B 0.94 &  &  &  &  \\*
10 & & 11.97 & 0.74 & 10.67 & 0.58 & 9.39 & 0.61 & 8.24 & 0.91 & 6.29 & 0.91 & 3.96 & 0.94 \\*
20 & & 18.26 & 0.66 & 10.04 & 0.52 & 7.65 & 0.58 & 17.82 & 0.62 & 11.93 & 0.64 & 7.57 & 0.71 \\*
50 & & 45.80 & -0.10 & 14.47 & 0.22 & 10.44 & 0.25 & 23.93 & -0.21 & 39.64 & -0.01 & 36.47 & -0.14 \\

\toprule
\multicolumn{1}{c}{} & \multicolumn{1}{c}{} & \multicolumn{12}{c}{Dataset} \\*
\cmidrule{3-14}

\multirow{3}{*}{\begin{tabular}[c]{@{}c@{}}Lost \\(\%)\end{tabular}}
& \multirow{3}{*}{\begin{tabular}[c]{@{}c@{}}rPPG \\Meth.\end{tabular}}
& \multicolumn{6}{c}{UCLA-rPPG}
& \multicolumn{6}{c}{UBFC-Phys} \\*
\cmidrule(lr){3-8} \cmidrule(lr){9-14}

& 
& \multicolumn{2}{c}{Strategy 0} & \multicolumn{2}{c}{Strategy 1} & \multicolumn{2}{c}{Strategy 2}
& \multicolumn{2}{c}{Strategy 0} & \multicolumn{2}{c}{Strategy 1} & \multicolumn{2}{c}{Strategy 2}\\*
\cmidrule(lr){3-4} \cmidrule(lr){5-6} \cmidrule(lr){7-8} \cmidrule(lr){9-10} \cmidrule(lr){11-12} \cmidrule(lr){13-14} 

& 
& \multicolumn{1}{c}{MAE} & \multicolumn{1}{c}{PCC} 
& \multicolumn{1}{c}{MAE} & \multicolumn{1}{c}{PCC} 
& \multicolumn{1}{c}{MAE} & \multicolumn{1}{c}{PCC} 
& \multicolumn{1}{c}{MAE} & \multicolumn{1}{c}{PCC} 
& \multicolumn{1}{c}{MAE} & \multicolumn{1}{c}{PCC} 
& \multicolumn{1}{c}{MAE} & \multicolumn{1}{c}{PCC} \\
\midrule

\multicolumn{1}{c}{} &  \multicolumn{1}{c}{\textcolor{green}{\textbf{NLM}}}& \multicolumn{12}{c}{} \\
\midrule

\B 0 & \multirow{4}{*}{OMIT} 
& \B 3.65 & \B 0.56 &  &  &  &  & \B 13.30 & \B 0.27 &  &  &  &  \\*
10 & & 10.47 & 0.60 & 4.87 & 0.55 & 3.45 & 0.60 & 18.90 & 0.25 & 13.28 & 0.27 & 13.30 & 0.27 \\*
20 & & 19.59 & 0.56 & 5.69 & 0.51 & 3.55 & 0.59 & 27.22 & 0.24 & 13.56 & 0.27 & 13.29 & 0.27 \\*
50 & & 58.04 & 0.21 & 9.70 & 0.41 & 3.38 & 0.66 & 56.45 & 0.07 & 15.01 & 0.21 & 13.11 & 0.28 \\
\midrule

\B 0 & \multirow{4}{*}{CHROM} 
& \B 3.76 & \B 0.59 &  &  &  &  & \B 13.65 & \B 0.25 &  &  &  &  \\*
10 & & 10.76 & 0.59 & 4.93 & 0.58 & 3.72 & 0.60 & 19.17 & 0.24 & 13.72 & 0.24 & 13.62 & 0.25 \\*
20 & & 19.82 & 0.53 & 5.84 & 0.52 & 3.83 & 0.59 & 27.11 & 0.24 & 13.61 & 0.26 & 13.56 & 0.25 \\*
50 & & 57.28 & 0.14 & 10.14 & 0.41 & 3.79 & 0.61 & 55.41 & 0.07 & 14.70 & 0.23 & 13.38 & 0.26 \\
\midrule

\B 0 & \multirow{4}{*}{POS} 
& \B 3.06 & \B 0.65 &  &  &  &  & \B 13.69 & \B 0.25 &  &  &  &  \\
10 & & 10.26 & 0.65 & 4.47 & 0.59 & 3.24 & 0.63 & 19.59 & 0.23 & 13.70 & 0.24 & 13.67 & 0.25 \\*
20 & & 19.79 & 0.60 & 5.59 & 0.52 & 3.33 & 0.61 & 28.20 & 0.23 & 13.64 & 0.25 & 13.62 & 0.25 \\*
50 & & 60.19 & 0.32 & 9.32 & 0.45 & 3.36 & 0.64 & 59.58 & 0.09 & 14.89 & 0.22 & 13.47 & 0.26 \\
\midrule

\multicolumn{1}{c}{} &  \multicolumn{1}{c}{\textcolor{orange}{\textbf{DLM}}}& \multicolumn{12}{c}{} \\
\midrule

\B 0 & \multirow{4}{*}{\begin{tabular}[c]{@{}c@{}}Efficient\\Phys\end{tabular}} 
& \B 7.99 & \B 0.37 &  &  &  &  & \B 15.93 & \B 0.17 &  &  &  &  \\
10 & & 12.45 & 0.38 & 7.78 & 0.41 & 7.40 & 0.38 & 19.42 & 0.18 & 16.34 & 0.16 & 15.77 & 0.17 \\*
20 & & 19.80 & 0.29 & 8.71 & 0.36 & 6.75 & 0.45 & 24.50 & 0.13 & 16.25 & 0.16 & 15.50 & 0.19 \\*
50 & & 39.79 & 0.00 & 11.16 & 0.33 & 6.41 & 0.47 & 37.64 & 0.00 & 17.32 & 0.12 & 15.31 & 0.17 \\
\midrule
 
\B 0 & \multirow{4}{*}{\begin{tabular}[c]{@{}c@{}}Contrast\\Phys\end{tabular}} 
& \B 7.13 & \B 0.40 &  &  &  &  & \B 16.87 & \B 0.17 &  &  &  &  \\*
10 & & 12.66 & 0.54 & 6.22 & 0.52 & 5.20 & 0.55 & 21.61 & 0.24 & 14.44 & 0.23 & 14.21 & 0.25 \\*
20 & & 20.00 & 0.61 & 5.73 & 0.61 & 4.29 & 0.65 & 28.42 & 0.30 & 12.79 & 0.30 & 12.56 & 0.31 \\*
50 & & 37.13 & -0.14 & 15.95 & 0.02 & 11.89 & 0.03 & 40.92 & -0.01 & 19.11 & 0.03 & 17.55 & 0.04 \\
\bottomrule

\addlinespace
\multicolumn{14}{l}{\parbox{16cm}{\footnotesize\ The bold denotes the baseline result (BS) for images sized 72x72 pixels with 0\% lost sample.}}

\end{tabular}
}
\end{table*}
%\twocolumn

DLM methods generally handle datasets with missing frames more effectively than NLM methods across most datasets. However, after implementing mitigation strategies, NLM methods exhibit significant improvements, even outperforming DLM methods in several datasets.
 
Furthermore, it is evident that the second mitigation strategy, involving signal reconstruction through linear interpolation, yields superior results compared to the first strategy of recalculating a new FPS. Nonetheless, it is important to note that in scenarios where information loss is substantial, and there isn't sufficient data for the second strategy, the results from the first strategy still show considerable improvement compared to having no mitigation strategy at all.

\subsection{Artifact Influence on Face Extraction}
In real-world settings, extracting faces from frames with artifacts is challenging, often leading to missed extractions. Our research introduces artifacts after face extraction to minimize instability caused by missed faces. Although we focus on how artifacts affect heart rate estimation, we conduct an additional experiment to explore their impact on face extraction. Using the LGI-PPGI dataset, we visually simulate real-world artifact scenarios to assess face extraction tools such as Mediapipe. The insights contribute to understanding challenges in face extraction amid artifacts.

\begin{figure*}[ht!]
  \begin{center}
    \includegraphics[width=\textwidth]{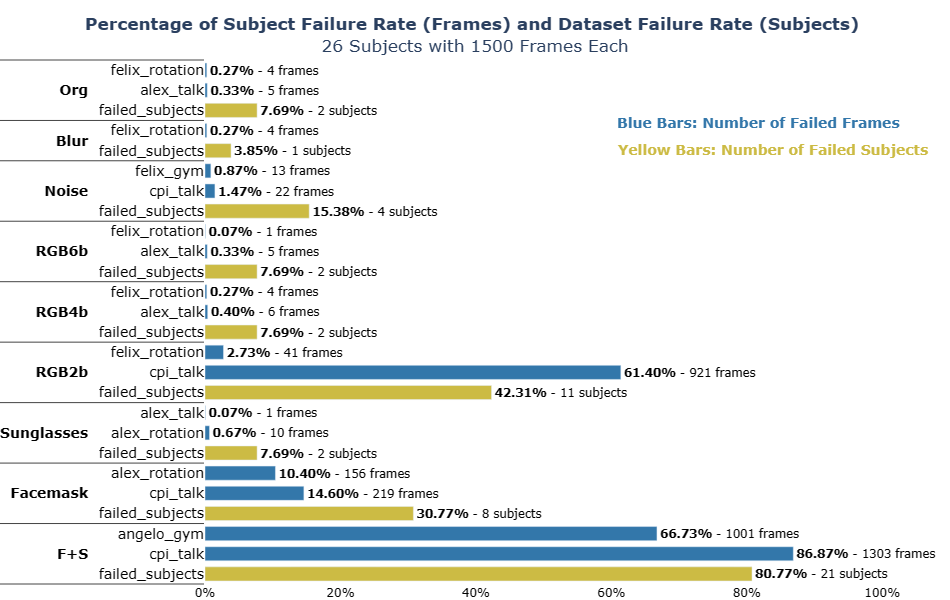}
  \end{center}
  \caption{Artifact Influence on LGI-PPGI Dataset}
  \label{fig:detail}
\end{figure*}

As observed in the yellow bars "failed\_subjects" in Fig~\ref{fig:detail}, most of the artifacts and color bit reductions introduced in this research did not significantly impact face extraction in subjects, except in three simulated scenarios: adding facemasks and sunglasses ranked first, RGB 2 bits second, and adding facemask to the face ranked last. Among these scenarios, the most impactful was the addition of facemasks and sunglasses, which failed in extracting faces for 21 out of 26 subjects, although not all frames in failed subjects experienced extraction failure.

%%% FIG

Further analysis focuses on the number of frames that failed to be extracted in specific subjects, as shown blue bar in Fig~\ref{fig:detail}. Notably, adding blur to datasets did not significantly affect Mediapipe face extraction, and in some cases, it even improved detection, as seen in the "alex\_talk" subject. Surprisingly, reducing the original RGB 8 bits to 6 bits yielded a similar improvement in the "felix\_rotation" subject. A trend in failed face extraction across transformation types is observed, with occurrences being most common in subjects "alex", "cpi", and "felix", particularly during specific tasks such as talking, rotation, and gym activities.

Although face extraction failures were not as pronounced as one might expect, certain scenarios, like "angelo\_gym" and "cpi\_talk", exhibited face extraction failures exceeding 50\%, which could significantly impact heart rate estimation. Mitigation strategies, such as using previously (or later) detected landmarks to extract faces or employing random missing frames mitigation strategies as those proposed before, can help mitigate the impact of failed extractions. However, it is important to note that these strategies are only effective if accurate face detection is possible.

    \label{sec:result}
    
\section{Conclusion}
    This article highlighted that rPPG extraction in real-world conditions, is impacted by intricate challenges posed by various artifacts, including spatial, temporal, and visual occlusions, that need a comprehensive evaluation approach. Through realistic simulation and comparative experimentation, we could observe nuanced insights into the effects of these factors on the performance of heart rate estimation from video streams. 

Regarding face extraction under artifacts, our findings indicate that most introduced artifacts and color bit reductions had negligible effects on face extraction in subjects. However, specific scenarios, such as the addition of facemasks and sunglasses, proved more challenging, leading to extraction failures in a limited number of datasets. Although face extraction failures were not as bad as maybe expected, our analysis underscores the importance of employing mitigation strategies such as utilizing previously detected landmarks or careful timestamping of the frames to reconstruct signals with a significant amount of randomly missing samples.

The resolution of the facial regions were found to have minimal impact on heart rate estimation, with certain methods even showing slight improvements at specific sizes, maybe due to the low-pass filtering effects cause by the downsizing of the regions. Notably, the choice of input size for comparison emerged as a crucial consideration, especially when dealing with deep learning methods that require very specific input formats. 
%Additionally, normal facial movements and facial symmetry had minimal impact on heart rate estimation accuracy, underscoring the robustness of the rPPG methodology under real-world conditions. 
The reduction in bit depth consistently influenced heart rate estimation performance, with lower bit depths leading to lower performance especially for heavy reductions. On the other hand, the impact of image degradation effects such as blur and noise varied across datasets and methods. Among those blur had a mild effect and occasionally even improved performance, maybe also due to the spurious low pass filtering effect caused by averaging neighbouring pixels. Comparatively, noise consistently exerted a more pronounced adverse effect, particularly on non-learning methods. Although denoising techniques showed promise in improving image quality, their corresponding impact on heart rate estimation performance was not really significant.

Visual occlusions posed challenges to heart rate estimation accuracy, with facemasks significantly impairing performance while sunglasses had minimal impact. Mitigation strategies varied in effectiveness. Deep learning methods showed considerable improvements in the facemask scenario, but the performance of a simple inpainting method such as removing the occluded skin regions showed better results in heart rate estimation, while more sophisticated inpainging methods such as GAN-based had superior performance in face visualization that did not translate to heart-rate estimation..

%Surprisingly, the combined presence of facemasks and sunglasses yielded unexpected outcomes, potentially due to the contrasting colors of the occluding objects

Furthermore, our analysis of frame rate variations revealed intriguing behavior in DLMs, with EfficientPhys demonstrating improved performance at lower frame rates in most datasets, except for UBFC-rPPG. ContrastPhys, a method that worked relatively well in all types of datasets due to its unsupervised nature, presented specific challenges at lower frame-rates. On the other hand, the results for NLMs remained relatively consistent with minor fluctuations across datasets. In the context of random sample missing scenarios, our experiment showed the comparative performance of NLMs and DLMs, with DLMs generally handling missing frames more effectively. However, mitigation strategies significantly improved the performance of NLMs, sometimes even beyond DLMs,  particularly through signal reconstruction using careful timestamping, which underscores the importance of being aware of potential data loss effects.

In conclusion, our work contributes to advancing the understanding of rPPG methodologies and provides practical insights for optimizing their performance in real-world applications. Future research directions may involve further refinement of our mitigation strategies and exploration of additional factors that influence the rPPG performance in diverse scenarios.
    \label{sec:conclusion}

\printcredits

\section*{Declaration of competing interest}
None

\section*{Acknowledgments}
We would like to express our sincere appreciation to the funding agencies that supported this study. We specifically acknowledge the Academy of Finland 6G Flagship program (Grant 346208) and PROFI5 HiDyn (Grant 32629) for their aid. We are also grateful to the Japan Society for the Promotion of Science (JSPS) for their support under KAKENHI Grant Number 21J22170.

%%%%%%%%%%%%%%%%%%%%
%%%%% APPENDIX %%%%%
%%%%%%%%%%%%%%%%%%%%

\appendix
\section*{Appendix A. Supplementary Data}
Supplementary data to this article can be found in other file attachments.

%\bibliographystyle{model1-num-names.bst}
%\bibliographystyle{cas-model2-names}
\bibliographystyle{elsarticle-num}

\bibliography{ms.bib}

\end{document}

\title{Supplemental Material - Evaluation of Video-Based rPPG in challenging environments: Artifact Mitigation and Network Resilience}  

\tnotetext[1]{This research has been supported by the Academy of Finland 6G Flagship program under Grant 346208 and PROFI5 HiDyn under Grant 32629 and JSPS (Japan Society for the Promotion of Science) KAKENHI Grant Number 21J22170.}

\author[1]{Nhi Nguyen}
\ead{thi.tn.nguyen@oulu.fi}

\author[1]{Le Nguyen}
\ead{le.nguyen@oulu.fi}

\author[1,2]{Honghan Li}
\ead{lihonghan@mbm.me.es.osaka-u.ac.jp}

\author[1,3]{Miguel \ Bordallo\ López}
\ead{miguel.bordallo@oulu.fi}

\author[1]{Constantino \'{A}lvarez\ Casado}
\ead{constantino.alvarezcasado@oulu.fi}

\affiliation[1]{organization={Center for Machine Vision and Signal Analysis (CMVS), University of Oulu},
            city={Oulu},
            country={Finland}}
            
\affiliation[2]{organization={Division of Bioengineering, Graduate School of Engineering Science, Osaka University},
            city={Osaka},
            country={Japan}}
\affiliation[3]{organization={VTT Technical Research Center of Finland Ltd.},
            city={Oulu},
            country={Finland}}

\begin{abstract}
This is the supplemental material for the article titled "Evaluation of Video-Based rPPG in Challenging Environments: Artifact Mitigation and Network Resilience". In this supplemental material we offer some alternative and complementary experiments and results that although they still offer insights about the extraction of rPPGs , they were deemed of smaller interest due to their relatively unsurprising results. In particular, the supplemental material discusses denoising strategies and also the combination of different occlusion methods such as glasses and facemasks combined.      
\end{abstract}

\maketitle

\setcounter{equation}{0}
\setcounter{figure}{0}
\setcounter{table}{0}
\setcounter{page}{1}
\setcounter{section}{0}
\setcounter{NAT@ctr}{0}

\makeatletter
\renewcommand{\theequation}{S\arabic{equation}}
\renewcommand{\thesection}{S\arabic{section}}
\renewcommand{\thefigure}{S\arabic{figure}}
\renewcommand{\thetable}{S\arabic{table}}
\renewcommand{\bibnumfmt}[1]{[S#1]}
\renewcommand{\citenumfont}[1]{S#1}
%\newcounter{SIfig}
%\renewcommand{\theSIfig}{S\arabic{SIfig}
%%%%%%%%%% Prefix a "S" to all equations, figures, tables and reset the counter %%%%%%%%%%

\section{Denoising}

\begin{figure*}[ht!]
  \begin{center}
    \includegraphics[width=\textwidth, height=0.3\textheight]{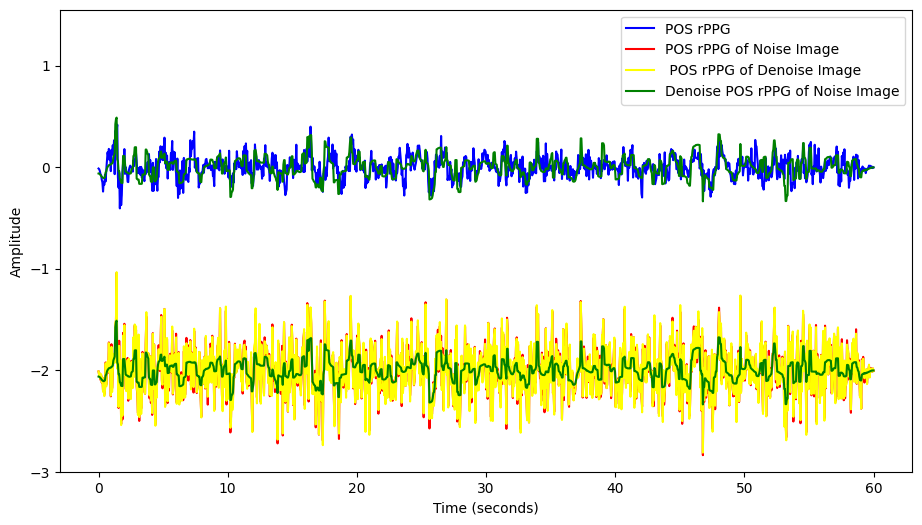}
  \end{center}
  \caption{Examples of four types of POS rPPG signals}
  \label{fig:denosing}
\end{figure*}

This section provides supplementary material for denoising experiments. In the denoising experiment in Section 6.4 of our article, it was found that while applying NAFNet and NLM improved image quality, it did not yield an enhancement in heart rate estimation from rPPG signals. To delve deeper into this phenomenon, a supplementary experiment is undertaken. This new experiment employs two distinct methods: one for denoising images (TVI), mirroring the previous approach, and another for denoising rPPG signals (TVS). Both methods utilize total variance denoising from the scikit-image library \cite{scikit-image}, integrating parameters such as $\epsilon$ set to 0.0002, 200 iterations, and a weight of 0.25. By employing this dual-pronged approach, the aim is to unravel potential differences in noise characteristics between images and rPPG signals, and how these disparities may affect the accuracy of heart rate estimation, shown in Table \ref{tab:deter_supp}.
%to gain further insights into the discrepancy observed between image quality enhancement and its impact on rPPG-based heart rate estimation. This endeavor seeks
From the observation of results, it is evident that TVI denoising performs similarly to the two denoising methods used in the previous experiment. However, when it comes to TVS denoising, there's a notable improvement in terms of MAE, although the PCC doesn't show significant improvement and sometimes even decreases. For a more detailed understanding of how rPPG signals are processed, Figure \ref{fig:denosing} illustrates examples of four types of POS rPPG. This visualization helps elucidate the relationship with the results table. Comparing the rPPG of denoised images and the rPPG of noisy images, there isn't much difference observed. However, the denoised rPPG of noisy images appears smoother, albeit at certain points it lacks peak information.

\begin{table*}[hb]
\captionsetup{justification = centering}
\centering
\caption{Heart Rate Estimation Errors: Baseline, Noise, and Denoising Datasets }
\label{tab:deter_supp}
\resizebox{0.9\textwidth}{!}{
\begin{tabular}{l *{15}{c}}
\toprule 
\multicolumn{1}{c}{} & \multicolumn{1}{c}{} & \multicolumn{14}{c}{Dataset} \\
\cmidrule{3-16}

\multirow{2}{*}{Trans.}
& \multirow{2}{*}{\begin{tabular}[c]{@{}c@{}}rPPG \\Meth.\end{tabular}}
& \multicolumn{2}{c}{COHFACE}
& \multicolumn{2}{c}{LGI-PPGI} 
& \multicolumn{2}{c}{MAHNOB}
& \multicolumn{2}{c}{PURE}
& \multicolumn{2}{c}{UBFC-rPPG}
& \multicolumn{2}{c}{UCLA-rPPG}
& \multicolumn{2}{c}{UBFC-Phys}    \\
\cmidrule(lr){3-4} \cmidrule(lr){5-6} \cmidrule(lr){7-8} \cmidrule(lr){9-10} \cmidrule(lr){11-12} \cmidrule(lr){13-14} 
\cmidrule(lr){15-16}

\multicolumn{1}{c}{}& \multicolumn{1}{c}{}
& \multicolumn{1}{c}{MAE} & \multicolumn{1}{c}{PCC} 
& \multicolumn{1}{c}{MAE} & \multicolumn{1}{c}{PCC}  
& \multicolumn{1}{c}{MAE} & \multicolumn{1}{c}{PCC}  
& \multicolumn{1}{c}{MAE} & \multicolumn{1}{c}{PCC}  
& \multicolumn{1}{c}{MAE} & \multicolumn{1}{c}{PCC} 
& \multicolumn{1}{c}{MAE} & \multicolumn{1}{c}{PCC} 
& \multicolumn{1}{c}{MAE} & \multicolumn{1}{c}{PCC}  \\
\midrule

\multicolumn{1}{c}{}  & \multicolumn{1}{c}{\textcolor{green}{\textbf{NLM}}} & \multicolumn{14}{c}{} \\
\midrule

BS & \multirow{4}{*}{OMIT} & 11.24 & 0.27 & 8.79 & 0.73 & 36.93 & 0.02 & 1.43 & 0.98 & 3.49 & 0.94 & 3.65 & 0.56 & 13.30 & 0.27 \\

\B Noise &  & \B 41.58 & \B 0.10 & \B 17.10 & \B 0.50 & \B 44.83 & \B 0.02 & \B 17.62 & \B 0.35 & \B 13.62 & \B 0.54 & \B 20.16 & \B 0.27 & \B 18.81 & \B 0.19 \\* 
TVI &  & 41.31 & 0.10 & \U{16.96} & \U{0.51} & \U{44.65} & \U{0.03} & 17.68 & 0.33 & 13.76 & 0.54 & \U{20.16} & \U{0.28} & 19.00 & 0.19 \\*
TVS &  & 15.28 & 0.08 & \U{13.86} & \U{0.53} & \U{33.81} & \U{0.04} & \U{8.20} & \U{0.62} & 12.24 & 0.54 & \U{7.33} & \U{0.43} & \U{14.44} & \U{0.20} \\
\midrule

BS & \multirow{4}{*}{CHROM} & 12.39 & 0.21 & 10.87 & 0.60 & 37.13 & -0.02 & 1.42 & 0.98 & 3.05 & 0.97 & 3.76 & 0.59 & 13.65 & 0.25 \\

\B Noise &  & \B 48.04 & \B 0.09 & \B 19.90 & \B 0.42 & \B 46.24 & \B 0.02 & \B 19.58 & \B 0.33 & \B 17.12 & \B 0.44 & \B 19.85 & \B 0.28 & \B 19.49 & \B 0.20 \\* 
TVI &  & 47.77 & 0.09 & 19.67 & 0.42 & \U{45.52} & \U{0.03} & 19.45 & 0.32 & 17.72 & 0.43 & 20.07 & 0.28 & 19.52 & 0.20 \\*
TVS &  & \U{37.09} & \U{0.10} & 18.89 & 0.40 & 43.82 & 0.02 & \U{16.73} & \U{0.35} & \U{15.70} & \U{0.45} & 16.74 & 0.28 & 18.64 & 0.19 \\
\midrule

BS & \multirow{4}{*}{POS} & 11.58 & 0.27 & 6.05 & 0.84 & 36.55 & 0.03 & 1.25 & 0.99 & 2.74 & 0.99 & 3.06 & 0.65 & 13.69 & 0.25 \\

\B Noise &  & \B 42.75 & \B 0.10 & \B 16.50 & \B 0.54 & \B 44.55 & \B 0.03 & \B 18.30 & \B 0.35 & \B 13.00 & \B 0.53 & \B 21.44 & \B 0.27 & \B 19.67 & \B 0.20 \\* 
TVI &  & 42.71 & 0.08 & 16.61 & 0.52 & 44.38 & 0.03 & 18.49 & 0.34 & \U{12.91} & \U{0.55} & 21.00 & 0.27 & 19.95 & 0.19 \\*
TVS &  & \U{16.91} & \U{0.13} & \U{13.80} & \U{0.55} & 37.52 & 0.01 & 13.02 & 0.35 & 13.24 & 0.49 & \U{10.72} & \U{0.28} & 15.47 & 0.18 \\
\midrule

\multicolumn{1}{c}{}  & \multicolumn{1}{c}{\textcolor{orange}{\textbf{DLM}}}& \multicolumn{14}{c}{} \\
\midrule
BS & \multirow{4}{*}{\begin{tabular}[c]{@{}c@{}}Efficient\\Phys\end{tabular}} & 17.46 & 0.06 & 18.33 & 0.37 & 38.35 & -0.03 & 10.12 & 0.50 & 7.73 & 0.69 & 7.99 & 0.37 & 15.93 & 0.17 \\ 

\B Noise &  & \B 23.10 & \B 0.00 & \B 27.27 & \B 0.08 & \B 40.04 & \B -0.04 & \B 27.07 & \B -0.01 & \B 32.81 & \B -0.02 & \B 26.06 & \B 0.05 & \B 23.60 & \B -0.01 \\* 
TVI &  & 21.35 & -0.01 & \U{25.36} & \U{0.19} & 40.08 & -0.03 & \U{23.07} & \U{0.06} & \U{30.80} & \U{0.03} & 20.83 & -0.03 & \U{22.55} & \U{0.01} \\*
TVS &  & 17.35 & -0.02 & \U{24.97} & \U{0.11} & \U{38.92} & \U{0.00} & \U{21.21} & \U{0.01} & 32.10 & -0.02 & 20.50 & 0.03 & 22.05 & -0.02 \\
\midrule

BS & \multirow{4}{*}{\begin{tabular}[c]{@{}c@{}}Contrast\\Phys\end{tabular}} & 9.67 & 0.22 & 13.86 & 0.50 & 46.52 & 0.06 & 11.63 & 0.59 & 3.72 & 0.94 & 7.13 & 0.40 & 16.87 & 0.17 \\

\B Noise &  & \B 10.10 & \B 0.10 & \B 14.06 & \B 0.51 & \B 46.58 & \B 0.08 & \B 26.31 & \B 0.16 & \B 6.11 & \B 0.81 & \B 15.11 & \B 0.17 & \B 21.33 & \B 0.07 \\* 
TVI &  & 10.43 & 0.13 & 14.51 & 0.50 & 46.54 & 0.07 & \U{25.73} & \U{0.19} & 6.63 & 0.79 & 15.24 & 0.13 & 22.44 & 0.06 \\*
TVS &  & 10.15 & 0.10 & 14.46 & 0.48 & 46.43 & 0.08 & 25.55 & 0.16 & 6.25 & 0.81 & \U{14.14} & \U{0.20} & \U{20.39} & \U{0.09} \\
\bottomrule
\addlinespace

\multicolumn{15}{l}{\parbox{18cm}{\footnotesize\ The term BS represents the normal image at 72x72 pixels, the bold represents the deteriorated result needing improvement, while the underline marks enhancements in both metrics compared to the bold.}}
\end{tabular}
}
\end{table*}

\section{Visual Occlusion and Mitigation}

\begin{figure*}[!ht]
    \centering
    \begin{subfigure}{\textwidth}
        \centering
        \includegraphics[width=\columnwidth, height=0.25\textheight, keepaspectratio]{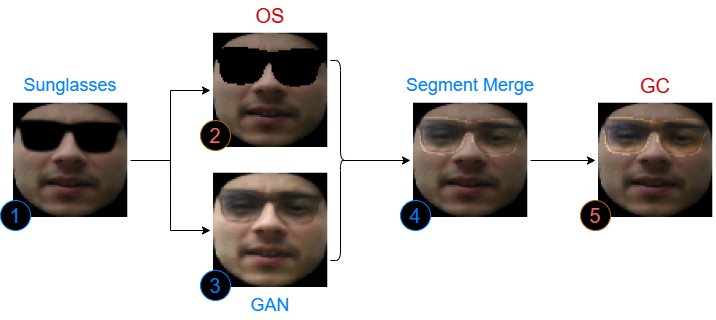}
        %\label{fig:sunglasses}
    \end{subfigure}
    \begin{subfigure}{\textwidth}
        \centering
        \includegraphics[width=\columnwidth, height=0.25\textheight, keepaspectratio]{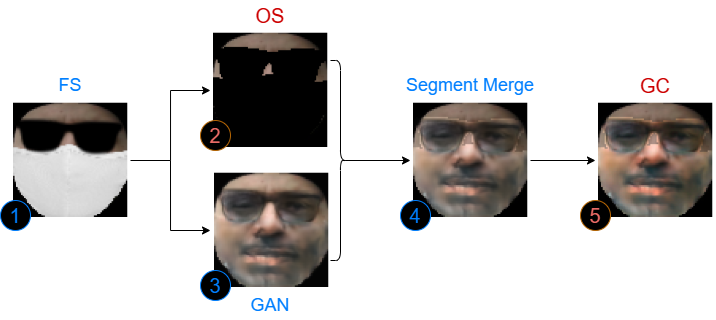}
        %\label{fig:both}
    \end{subfigure}
    \caption{Illustrations demonstrating sunglasses and both visual occlusion mitigation. 1) Occluded image, 2) Original skin image, 3) GAN generated image, 4) Merged image of original and generated skin, 5) Merged image with color transfer from real to generated skin.}
    \label{fig:supp_vo}
\end{figure*}

This section provides supplementary information regarding visual occlusion mitigation involving sunglasses and the combination of sunglasses and facemasks. Figures \ref{fig:supp_vo} respectively illustrate each stage of the process applied to facial images, demonstrating the mitigation methods for occlusions involving various types of eye and face combinations.

For heart rate estimation, as illustrated in Table~\ref{tab:ip_both}, the combined presence of facemasks and sunglasses surprisingly did not significantly impact the accuracy of heart rate estimation as anticipated. In fact, some results even showed improvement compared to those obtained with the facemask alone. This unexpected outcome could potentially be attributed to the contrasting colors of the mask and sunglasses, although not purely black or white, still interact in a complementary manner within a certain range.

\begin{table*}[ht]
\captionsetup{justification = centering}
\centering
\caption{Heart Rate Estimation Errors: Baseline, Occlusion, and Mitigating Datasets }
\label{tab:ip_both}
\resizebox{0.9\textwidth}{!}{
\begin{tabular}{l *{15}{c}}
\toprule 
\multicolumn{1}{c}{} & \multicolumn{1}{c}{} & \multicolumn{14}{c}{Dataset} \\
\cmidrule{3-16}

\multirow{2}{*}{Trans.}
& \multirow{2}{*}{\begin{tabular}[c]{@{}c@{}}rPPG \\Meth.\end{tabular}}
& \multicolumn{2}{c}{COHFACE}
& \multicolumn{2}{c}{LGI-PPGI} 
& \multicolumn{2}{c}{MAHNOB}
& \multicolumn{2}{c}{PURE}
& \multicolumn{2}{c}{UBFC-rPPG}
& \multicolumn{2}{c}{UCLA-rPPG}
& \multicolumn{2}{c}{UBFC-Phys}    \\
\cmidrule(lr){3-4} \cmidrule(lr){5-6} \cmidrule(lr){7-8} \cmidrule(lr){9-10} \cmidrule(lr){11-12} \cmidrule(lr){13-14} 
\cmidrule(lr){15-16}

\multicolumn{1}{c}{}& \multicolumn{1}{c}{}
& \multicolumn{1}{c}{MAE} & \multicolumn{1}{c}{PCC} 
& \multicolumn{1}{c}{MAE} & \multicolumn{1}{c}{PCC}  
& \multicolumn{1}{c}{MAE} & \multicolumn{1}{c}{PCC}  
& \multicolumn{1}{c}{MAE} & \multicolumn{1}{c}{PCC}  
& \multicolumn{1}{c}{MAE} & \multicolumn{1}{c}{PCC} 
& \multicolumn{1}{c}{MAE} & \multicolumn{1}{c}{PCC} 
& \multicolumn{1}{c}{MAE} & \multicolumn{1}{c}{PCC}  \\
\midrule

\multicolumn{1}{c}{}  & \multicolumn{1}{c}{\textcolor{green}{\textbf{NLM}}} & \multicolumn{14}{c}{} \\
\midrule

BS & \multirow{4}{*}{OMIT} & 11.24 & 0.27 & 8.79 & 0.73 & 36.93 & 0.02 & 1.43 & 0.98 & 3.49 & 0.94 & 3.65 & 0.56 & 13.30 & 0.27 \\

\B FS &  & \B 19.76 & \B 0.08 & \B 9.37 & \B 0.76 & \B 36.72 & \B 0.07 & \B 2.83 & \B 0.89 & \B 11.74 & \B 0.42 & \B 6.46 & \B 0.43 & \B 15.33 & \B 0.21 \\* 
GC &  & 20.66 & 0.11 & 10.40 & 0.70 & 36.80 & 0.10 & 8.04 & 0.59 & \U{9.80} & \U{0.51} & 9.38 & 0.36 & 14.85 & 0.20 \\*
OS &  & \U{17.10} & \U{0.14} & \U{7.71} & \U{0.79} & \U{36.30} & \U{0.08} & \U{2.22} & \U{0.93} & \U{5.22} & \U{0.73} & 6.28 & 0.41 & \U{14.13} & \U{0.24} \\
\midrule

BS & \multirow{4}{*}{CHROM} & 12.39 & 0.21 & 10.87 & 0.60 & 37.13 & -0.02 & 1.42 & 0.98 & 3.05 & 0.97 & 3.76 & 0.59 & 13.65 & 0.25 \\

\B FS &  & \B 18.32 & \B 0.09 & \B 9.40 & \B 0.73 & \B 36.94 & \B 0.06 & \B 2.79 & \B 0.90 & \B 12.31 & \B 0.44 & \B 7.28 & \B 0.42 & \B 15.07 & \B 0.22 \\* 
GC &  & 20.10 & 0.09 & 13.08 & 0.59 & 37.30 & 0.06 & 8.57 & 0.61 & \U{9.58} & \U{0.58} & 10.44 & 0.33 & \U{14.63} & \U{0.23} \\*
OS &  & \U{15.67} & \U{0.11} & 9.35 & 0.72 & \U{36.29} & \U{0.07} & \U{1.96} & \U{0.95} & \U{5.77} & \U{0.76} & \U{6.78} & \U{0.45} & \U{14.10} & \U{0.24} \\
\midrule

BS & \multirow{4}{*}{POS} & 11.58 & 0.27 & 6.05 & 0.84 & 36.55 & 0.03 & 1.25 & 0.99 & 2.74 & 0.99 & 3.06 & 0.65 & 13.69 & 0.25 \\

\B FS &  & \B 17.63 & \B 0.12 & \B 6.80 & \B 0.84 & \B 36.08 & \B 0.07 & \B 2.04 & \B 0.95 & \B 5.10 & \B 0.82 & \B 6.09 & \B 0.43 & \B 14.39 & \B 0.23 \\* 
GC &  & 21.61 & 0.09 & 9.79 & 0.70 & 36.82 & 0.10 & 8.47 & 0.58 & 6.63 & 0.72 & 9.49 & 0.34 & 14.64 & 0.23 \\*
OS &  & \U{16.66} & \U{0.13} & 7.42 & 0.78 & 35.98 & 0.07 & 2.23 & 0.93 & \U{3.30} & \U{0.95} & 6.10 & 0.45 & \U{14.01} & \U{0.25} \\
\midrule

\multicolumn{1}{c}{}  & \multicolumn{1}{c}{\textcolor{orange}{\textbf{DLM}}}& \multicolumn{14}{c}{} \\
\midrule
BS & \multirow{4}{*}{\begin{tabular}[c]{@{}c@{}}Efficient\\Phys\end{tabular}} & 17.46 & 0.06 & 18.33 & 0.37 & 38.35 & -0.03 & 10.12 & 0.50 & 7.73 & 0.69 & 7.99 & 0.37 & 15.93 & 0.17 \\ 

\B FS &  & \B 22.14 & \B 0.00 & \B 27.78 & \B 0.04 & \B 38.56 & \B -0.01 & \B 22.03 & \B 0.07 & \B 29.31 & \B 0.10 & \B 21.65 & \B 0.00 & \B 22.93 & \B 0.02 \\* 
GC &  & 21.56 & -0.02 & \U{24.50} & \U{0.12} & 38.03 & -0.01 & 24.84 & 0.01 & 31.11 & 0.02 & \U{21.53} & \U{0.04} & 23.42 & 0.01 \\*
OS &  & 47.86 & 0.01 & 62.17 & 0.10 & 51.81 & -0.07 & 105.76 & -0.05 & 71.21 & -0.02 & 104.76 & 0.01 & 103.32 & 0.01 \\*
\midrule

BS & \multirow{4}{*}{\begin{tabular}[c]{@{}c@{}}Contrast\\Phys\end{tabular}} & 9.67 & 0.22 & 13.86 & 0.50 & 46.52 & 0.06 & 11.63 & 0.59 & 3.72 & 0.94 & 7.13 & 0.40 & 16.87 & 0.17 \\

\B FS &  & \B 12.76 & \B 0.03 & \B 18.36 & \B 0.28 & \B 46.95 & \B 0.04 & \B 30.92 & \B 0.00 & \B 12.50 & \B 0.44 & \B 21.60 & \B 0.05 & \B 24.07 & \B 0.08 \\* 
GC &  & 11.60 & 0.00 & 18.59 & 0.29 & 47.34 & 0.08 & 31.06 & 0.17 & 12.74 & 0.40 & 22.88 & 0.08 & 28.36 & 0.03 \\*
OS &  & \U{11.42} & \U{0.06} & 19.89 & 0.21 & \U{46.51} & \U{0.08} & \U{29.35} & \U{0.11} & 14.76 & 0.28 & 22.23 & 0.03 & 27.97 & 0.05 \\
\bottomrule
\addlinespace

\multicolumn{15}{l}{\parbox{18cm}{\footnotesize\ The term "BS" stands as the baseline result for images sized 72x72 pixels without occlusion, the bold represents the result of visual occlusion, while the underline marks enhancements in both metrics compared to the bold. "FS" indicates the presence of facemask and sunglasses. "OS" refers to the original skin region method, while "GC" represents GAN-OS with the color transfer method.}}

\end{tabular}
}
\end{table*}

In terms of the mitigation strategy for heart rate estimation with NLM, in cases of facemasks and sunglasses combined, the GAN method exhibited improvement in a few datasets, while the rest performed worse than the occluded images. Surprisingly, the original skin region approach with little skin region remains demonstrated a slight improvement in some datasets and methods. Regarding the mitigated strategy for heart rate estimation with DLM, for facemasks, the two DLM methods demonstrated different trends. While ContrastPhys exhibited a slight improvement in some datasets with both mitigate methods, EfficientPhys performed poorly and only showed a very slight increase in some datasets with the GANs method.

\bibliographystyle{elsarticle-num}
\bibliography{supplement.bib}